%% file: pattern_retrieval.tex
\documentclass[3p]{elsarticle}

\usepackage{lineno,hyperref}
\usepackage{amsmath}
\usepackage{amssymb}
\usepackage{algorithm, algpseudocode}
\usepackage{mathtools}
\usepackage{float}
\usepackage{pgfplots}
\usepackage{graphicx}
\usepackage{array}
\usepackage{enumitem}
\usepackage{multirow}
\usepackage{arydshln}
\usepackage{subcaption}

\usepgfplotslibrary{patchplots}
\usepgfplotslibrary{fillbetween}
\pgfplotsset{compat=1.11}

\modulolinenumbers[100]

\newcommand{\argmax}{\mathop{\mathrm{arg\,max}}}
\newcolumntype{L}[1]{>{\raggedright\let\newline\\\arraybackslash\hspace{0pt}}m{#1}}
\newcolumntype{C}[1]{>{\centering\let\newline\\\arraybackslash\hspace{0pt}}m{#1}}
\newcolumntype{R}[1]{>{\raggedleft\let\newline\\\arraybackslash\hspace{0pt}}m{#1}}

\newtheorem{definition}{Definition}

\journal{...}

\biboptions{authoryear}

\begin{document}
	
	\begin{frontmatter}
		
		\title{Pattern retrieval of traffic congestion using graph-based associations of traffic domain-specific features}
		
		\author{Tin T. Nguyen\corref{cor1}}
		\author{Simeon C. Calvert}
		\author{Guopeng Li}
		\author{Hans van Lint}
		
		\address{Department of Transport and Planning, Delft University of Technology, The Netherlands}
		%
		
		\begin{abstract}
			The fast-growing amount of traffic data brings many opportunities for revealing more insightful information about traffic dynamics. However, it also demands an effective database management system in which information retrieval is arguably an important feature. The ability to locate similar patterns in big datasets potentially paves the way for further valuable analyses in traffic management. This paper proposes a content-based retrieval system for spatiotemporal patterns of highway traffic congestion. There are two main components in our framework, namely pattern representation and similarity measurement. To effectively interpret retrieval outcomes, the paper proposes a graph-based approach (relation-graph) for the former component, in which fundamental traffic phenomena are encoded as nodes and their spatiotemporal relationships as edges. In the latter component, the similarities between congestion patterns are customizable with various aspects according to user expectations. We evaluated the proposed framework by applying it to a dataset of hundreds of patterns with various complexities (temporally and spatially). The example queries indicate the effectiveness of the proposed method, i.e. the obtained patterns present similar traffic phenomena as in the given examples. In addition, the success of the proposed approach directly derives a new opportunity for semantic retrieval, in which expected patterns are described by adopting the relation-graph notion to associate fundamental traffic phenomena.
			
		\end{abstract}
		
		
	\end{frontmatter}
	
	\linenumbers
	
	\section*{Highlights}
	\begin{itemize}
		\item[-] The vast amount of traffic data demands an effective database management system in which information retrieval has been shown to be highly valuable.
		\item[-] A systematic, automatic method for content-based (i.e. by giving a visual example)  highway-traffic congestion pattern retrieval is proposed.
		\item[-] Congestion patterns are dynamically represented as graphs in which fundamental traffic phenomena are encoded as nodes and their spatiotemporal relationships as edges.
		\item[-] The similarities between congestion patterns are customisable with various aspects according to user expectations.
		\item[-] A case study demonstrates the effectiveness of the proposed graph-based method in retrieving similar patterns to a given pattern.
	\end{itemize}
	
	\input{main_sections}
	
	
	\bibliography{bibfile}
	
\end{document}

%% file: main_sections.tex
\section{Introduction}
\subsection{The necessity of congestion pattern retrieval}
Innovation technologies have opened a new area of big data in many domains. On the one hand, these data provide many opportunities to gain valuable insights which are critically important for strengthening knowledge in any domain. On the other hand, vast amounts of data pose huge challenges in management and utilisation. In the field of transportation AVI - Automatic Vehicle Identification and FCD - Floating Car Data, many highly relevant traffic quantities, such as vehicular speeds or volumes, can now be collected in different degrees of spatial and temporal granularity, in part thanks to a variety of sensing systems such as induction loops. These data are beneficial for various purposes in road administration, industry and academia, including policy evaluation \citep{van2005accurate, wang2006renaissance, calvert2011modelling}, traffic management \citep{calvert2018improving}, traffic modelling and simulation \citep{van2018hierarchical, soriguera2011estimation, vlahogianni2005optimized}.

The collected traffic data comprise of critical information for understanding many aspects of traffic, none so important as traffic congestion as a major nuisance and of major economic influence. Traffic congestion can be triggered at many different times and places due to various different reasons such as increasing travel demands at peak hours or incidents on roads, but also minor behavioural disturbances. Hence, we can expect numerous instances of congestion to exist in historical traffic data, which constitutes a valuable source of insights into traffic congestion. Arguably, it is advantageous to have a retrieval system that can identify similar congestion instances in historical traffic data. Such a retrieval system can pave the way for the development of many applications, such as traffic analysis and traffic prediction to analyse the recurrence of congestion. By checking if similar types of congestion occurred in the past, we can evaluate if a given congestion instance is recurrent or non-recurrent. In addition, if a type of congestion is recurrent, (possible) variations can be revealed based on analysing similarly occurring instances. A further step in this direction is to associate these similar instances with other relevant sources of information, e.g. incidents or topology so that a more thorough understanding of certain types of congestion is derived. On the traffic prediction branch, the ability to extract similar patterns in historical data is a convenient tool for identifying (ir)regularities in traffic, which is highly valuable for predicting traffic states (e.g., \citep{lopez2017revealing}). To the best knowledge of the authors, such a retrieval system for traffic congestion is currently lacking in transportation literature. This gap hampers the potential benefits of the collected traffic data, in particular, to obtain similar instances of congestion for different purposes and aid traffic analysis and prediction in congested traffic.


The dynamics of traffic involve both the spatial and temporal dimensions. Hence, congestion is effectively observed or evaluated by constructing two-dimensional maps of relevant data like speeds or flows. Such representation is visually intuitive and also reveals relevant insights into traffic phenomena like wide-moving jams. This 2D representation of congestion motivates the creation of 2D maps, equivalently images, for traffic congestion occurrences, particularly at the corridor level. These maps are so-called congestion patterns. Subsequently, various techniques from computer vision will be applicable for different studies. For example, \cite{nguyen2019feature} applied image segmentation methods to detect the two most common traffic phenomena, namely traffic disturbances and homogeneous regions \citep{helbing2009theoretical}. The paper also shows that, compared to low-level image features \citep{bay2008speeded}, these traffic-specific elements can lead to more crisp clusters of congestion patterns. Since the constructed feature vector is a histogram of particular traffic regions, spatiotemporal relationships are not taken into account, which can hinder the performance when directly applying to pattern retrieval, in which (probably a limited number of) the most similar patterns are searched for from the dataset. Nevertheless, the benefits of representing congestion patterns as images motivate an approach focused on an image-like pattern retrieval system for congested traffic.

In general, there are two types of frameworks for image retrieval systems, namely text-based and content-based \citep{datta2008image}. In the former approach, each image is annotated with labels or so-called keywords when preparing a database. By specifying a particular keyword, the related patterns are easily identified and retrieved. The advance of this approach is that the retrieval mechanism is simple to implement. However, annotating images is usually done by hand, which is time-consuming and prone to errors due to numerous available items. Retrievals in the latter approach (see \citep{zhou2017recent} for a recent literature survey) are proceeded by proving an example image in advance. The retrieval system automatically extracts features from this image and searches for matches from available images in a database. Commonly used features are low-level information like colour, shape, and texture. This approach can lead to an uninterpretable connection between the low-level visual features and their conceptual meanings. Therefore, this drawback limits the interpretations of retrieval outcomes.

\begin{figure}
	\centering
	\begin{minipage}{0.8\textwidth}%
		\centering
		\qquad
		\includegraphics[width=\textwidth]{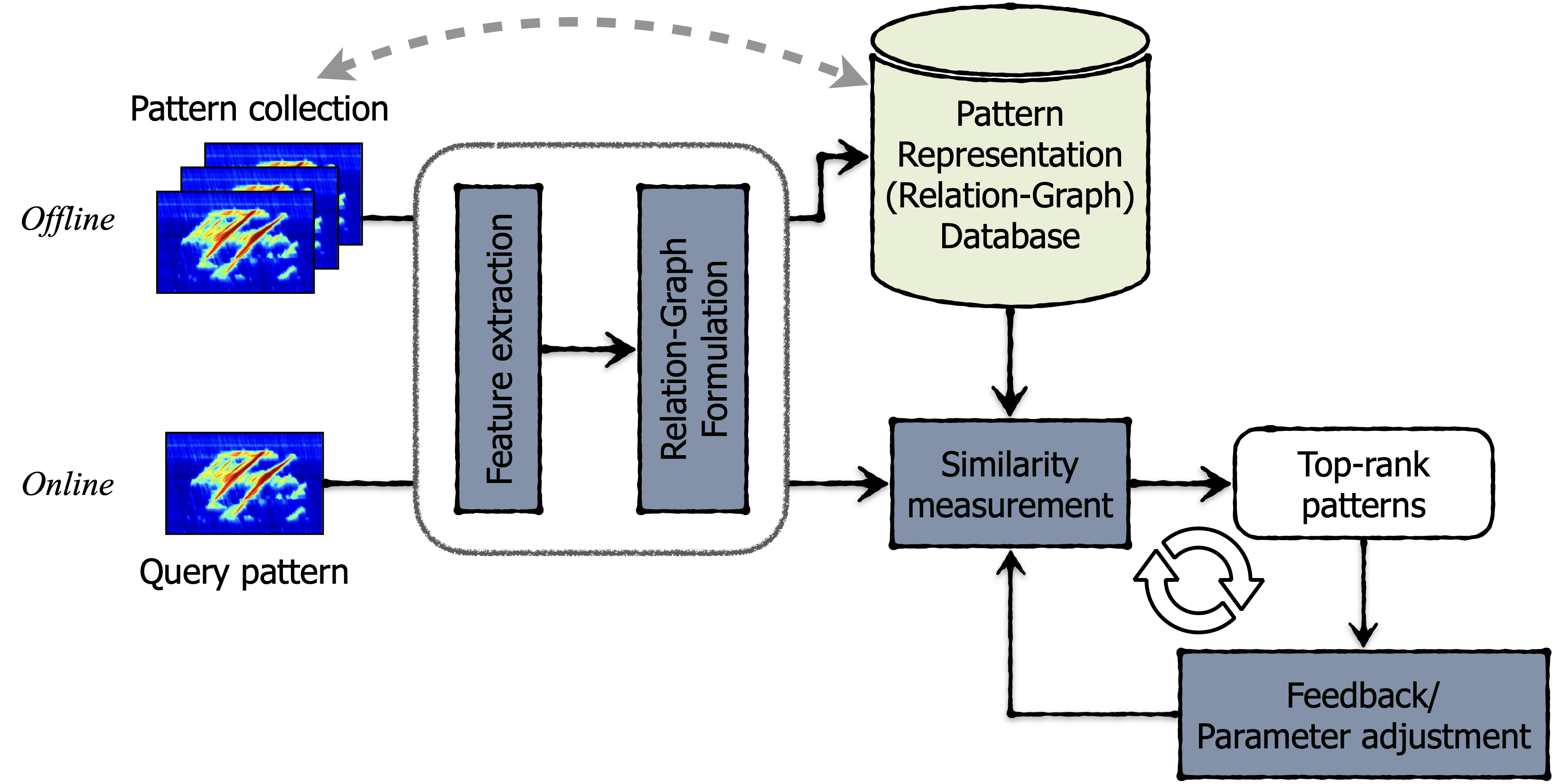}
		\caption{The overall framework for congestion pattern retrieval.}
		\label{fig:retrieval_framework}
	\end{minipage}%
\end{figure}

To this end, this paper proposes an information retrieval framework for occurrences of traffic congestion and simultaneously aims to achieve a high level of interpretability of retrieval outcomes. In particular, each entire occurrence is captured and represented by a two-dimensional speed map, which acts as a fundamental object in the framework. Distinctive representation of a pattern is constructed at an abstract level using a graph-based association of traffic-specific regions. Traffic-domain features are essential for obtaining an abstract description of congestion patterns. The application of graphs preserves (possible) relations between components in a pattern, which essentially illustrates the overall structure. Moreover, to measure the similarity between two patterns, a matching method is formulated and parameterised regarding several observable characteristics of patterns. Accordingly, various expectations of patterns obtained from a query can be intuitively translated into proper configurations of those parameters.

\subsection{General retrieval framework}
Fig \ref{fig:retrieval_framework} illustrates the overall framework of our proposed retrieval system. There are two major components which are the so-called pattern representation and similarity measurement. The former consists of two sub-components: the feature extraction and the relation-graph formulation. The feature extraction determines and extracts important traits from a pattern, from which patterns can be compared. Afterwards, these individual characteristics need to be integrated into a cohesive instance that represents patterns. In this paper, we propose employing graphs for that purpose due to their capability to preserve relations between regions in a pattern. The combination of these two sub-components leads to an abstract representation of any given congestion pattern. The similarity measurement component determines the degree of resemblance between two patterns. Its inputs include two relation graphs representing two patterns with its output being a similarity score.

In the offline stage, patterns are processed such that related features and their relation graphs are built and registered on a database. Patterns are retrieved in the online stage. If a retrieval outcome does not meet certain expectations, users can repeat the retrieval process to improve the result. There are two approaches for designing this iterative improvement, namely relevant feedback or parameter adjustment. In the former, users indicate whether individual obtained patterns are relevant or irrelevant. A proper method is implemented that takes users' feedback and revises the retrieval result accordingly. In the latter approach, the retrieval system is not automated. Based on their understanding of the retrieval framework, users can tune parameters to obtain desired patterns. Note that this last part is out of the scope of this paper.

\subsection{Paper outline}
The rest of the paper is organized as follows. Section \ref{sec_patternrepresentation} describes the process of extracting relevant features and constructing the so-called relation graphs as the representation for congestion patterns. The measurement of similarity between two patterns is presented in Section \ref{sec_similaritymeasure}. In Section \ref{sec_experimentandresult}, we describe an experiment for evaluating the proposed method. Finally, Section \ref{sec_conclusion} concludes this study.

\section{Pattern representation}
\label{sec_patternrepresentation}

To compare two patterns, we need to identify their representative characteristics, which we will refer to as \textit{features}. Furthermore, it is important to develop a suitable mechanism for combining these features into a coherent structure. This allows the similarity between patterns to be computed and assessed effectively. This section respectively discusses these components in detail.

The feature extraction component aims to extract features that best describe an item, i.e. a congestion pattern in our application. They need to be selected such that patterns are well differentiated so that similarities are well measured. On the one hand, since congestion patterns are represented as images, low-level attributes, such as colour, shape, and texture, can be extracted to describe related patterns. On the other hand, as these images represent traffic congestion, they possess traffic phenomena that are well-acknowledged in the field. Traffic-related characteristics can provide a high-level semantic approach in formulating representative features for congestion patterns. In this paper, we adopt the latter approach due to their ability to provide transparency and interpretation. Before going into the details, for the coherence throughout the paper, we first define some common terminologies.

\subsection{Terminology}
\paragraph*{Congestion patterns} A congestion pattern represents congested traffic on a road stretch (or corridor) over a certain temporal period. In essence, it is a two-dimensional matrix of traffic states (such as speed, flow or density) where each value pertains to a traffic state on a road section and time period. To obtain such discrete values over location and time, we use the Adaptive Smoothing Method \citep{treiber2002reconstructing, schreiter2010two} to map the (irregularly available) sensor data on equidistant grids. The result is equivalent to an (intensity) image, which is convenient for observing the traffic therein. In this paper, the term \textit{congestion pattern} refers to both the image representation and the underlying 2D matrix of traffic states.

\paragraph*{Region (Area)} A region (or an area) refers to a group of connected pixels in the equivalent image representation of a congestion pattern. The connection is either by the spatial or temporal dimension.

\paragraph*{Traffic primitive (component)} A traffic primitive (or component) refers to a region representing a specified traffic phenomenon, which will be introduced in the following sections.

\subsection{Feature extraction}
Congestion patterns can show instances of these widely acknowledged phenomena, including wide-moving jams, homogeneously heavy congestion, and traffic bottlenecks. These are visually observable in image patterns of congestion. The former two components are used and evaluated by \cite{nguyen2019feature}. By combining these two with the extent of congestion to formulate a feature vector, the authors derive different meaningful clusters of congestion patterns. The latter component is also a regular phenomenon of congestion as they are a common cause of traffic congestion. In this work, we incorporate and evaluate these three components as fundamental domain features of congestion.
\subsubsection{Abstract primitives}
\paragraph*{Traffic disturbances}
Disturbances occur regularly in traffic and can be visualised effectively by spatiotemporal maps or traffic (image) patterns. They can emerge from a bottleneck where approaching vehicles try to synchronise with slow traffic therein. These disturbances can propagate further upstream and form wide-moving jams. \cite{krishnakumari2017traffic} proposed and successfully applied the Active Shape Model to identify WMJs in image representation of traffic congestion. Since minor disturbances and wide-moving jams have similar shapes except for their spatial extents, this paper further employs this method to determine those in congestion patterns.

The Active Shape Model technique \citep{cootes1995active} describes a shape using a mean shape and its variations, which are obtained from a set of similar training shapes. Thus, given a new shape, the error of fitting the shape model to this shape can be used to identify or classify the shape. To obtain these shapes, pattern images are segmented using the Watershed transformation \citep{nguyen2019feature} into different traffic state regions. The boundaries of these regions are identified and clustered by the Active Shape Model for detecting traffic disturbances. We refer to the original paper \citep{nguyen2019feature} for further details.

\paragraph*{Homogeneous congestion}
Homogeneous congestion represents the spreading of congested traffic over space and time with consistently low vehicular speeds. They are normally associated with strong bottlenecks or accidents where the mismatch between traffic demand and local supply is significant. The regions associated with homogeneous congestion are referred to as the Demand-Supply element in \citep{nguyen2019feature}, in which the authors propose a simple condition to detect their existence. In this paper, we adopt texture analysis for the identification of regions of homogeneous congestion.

\cite{haralick1973textural} proposed deriving various texture features of an image using a grey-level co-occurrence matrix (GLCM). It enables the calculations of different statistics to quantify/represent texture characteristics of the related image. This method has been one of the most popular approaches in image texture representation. Some widely used features are energy, contrast, homogeneity, and entropy. 
Each number in the GLCM shows how frequently the related pair of intensities is present in the related image with respect to a pre-defined (two-dimensional) offset.


A preliminary analysis suggests that the energy feature is most promising for identifying homogeneous regions in congestion patterns. In fact, energy is a measure of the homogeneity of an image. It is defined by Equation \ref{eq:glcm_energy}. The number of grey levels in a homogeneous region is expected to be low, which shifts the whole distribution to a small group of $p_d(i,j)$ ($p_d(i,j)$ represents the frequency of having the co-occurrence of intensities $i$ and $j$ at a certain distance $d$). The more homogeneous a region is, the higher the energy feature gets.

\begin{equation}
	Energy = \sum_{i=1}^{N}\sum_{j=1}^{N}p_d(i,j)^2
	\label{eq:glcm_energy}
\end{equation}

\paragraph*{Traffic bottleneck}
Traffic bottleneck detection has a large body of literature that includes many approaches and methods. In this study, we adopt the framework in \citep{nguyen2021automated}, which identifies bottleneck location and activation time from speed maps representing traffic congestion. Furthermore, the method also extracts the boundaries of upstream congestion regions, which are beneficial for further analyses. In principle, congestion regions are identified by applying the active contour model without edges \citep{chan2001active} - a well-known image segmentation technique in computer vision. The model formulates congestion as foreground and free-flow regions as background in the corresponding segmentation problem. Bottleneck locations are detected by observing speed gradients along the direction of characteristic waves. Discontinuities (drops of speed at upstream) of traffic speeds are associated with possible bottleneck activations. Both primary and secondary bottlenecks can be identified successfully by this method. We refer to the original paper for a complete description of the framework.

\subsubsection{Primitive characteristics}
The objective of pattern retrieval is to localise patterns that best resemble a given pattern. The judgement of possible resemblances between two patterns can come from different perspectives of traffic observers. 

We define two overall perspectives on the matching two congestion patterns, namely pattern abstract and element detail. The former is concerned with the present as well as relations of traffic phenomena primitives as shown in traffic congestion patterns. For example, a pattern represents an activated bottleneck that causes congested traffic upstream, in which disturbances may have emerged and spilt back. The relevant traffic phenomena thus include the bottleneck and those disturbances --- together they form the high-level structure. The latter perspective examines patterns \textit{within the structure} by comparing details of different components, for example, the bottleneck severity and the frequency with which disturbances have emerged.


Regarding pattern abstract, we differentiate between structural integrity and primitive completeness. The former prioritises similarities in overall structure (size, area) as much as possible, and is tolerant of missing details in target patterns; whereas the latter focuses on target patterns having as much of all the constituent elements as possible regardless of their placements (locations, times) and the resulting overall structure.

With respect to element details, there is possibly an infinite number of characteristics that can be considered. To demonstrate the proposed framework, we therefore make a selection. Specifically, we consider the size of the patterns and the constituting elements, for which we look (i) at the relative proportions of elements in the related pattern, and (ii) the absolute size of elements. Large relative proportions imply having similar patterns regardless of the absolute size, while absolute size focuses on the actual sizes of patterns. As an example application, the former is preferable when looking for patterns with similar traffic phenomena, whereas the latter is more suitable when matching the consequences of congestion is important.

\subsection{Relation-graph formulation}
Feature representation combines attributes that are extracted from patterns in such a way that makes those patterns distinctive. In this paper, we employ so-called \textit{relation-graphs} as an alternative to a vector (a list) of features. 
The difference between this relation graph and a feature vector representation is that a graph representation can describe not just the list of relevant features (encoded in the nodes) but also their relationships (encoded in the links between those nodes). 

\begin{figure}
	\centering
	\begin{minipage}{0.9\textwidth}%
		\centering
		\begin{subfigure}[b]{0.45\textwidth}
			\centering
			\includegraphics[width=\textwidth]{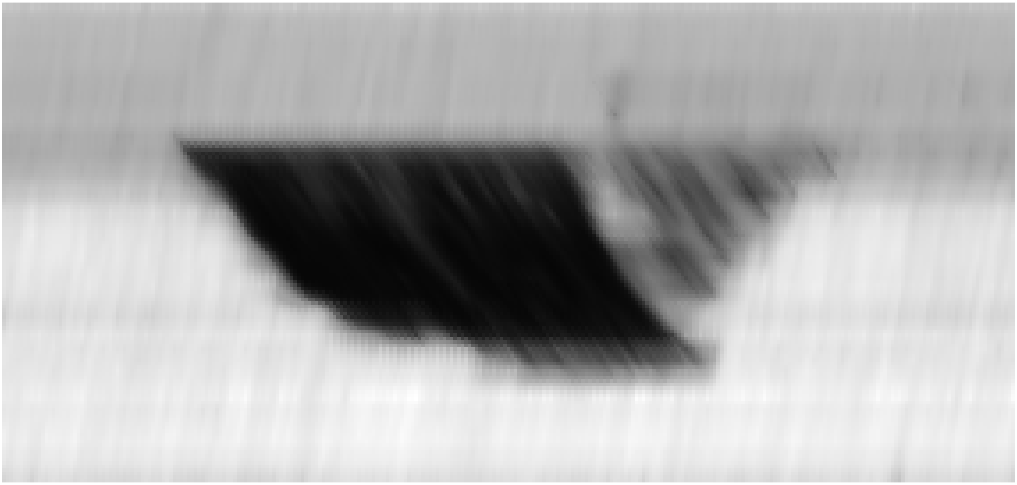}
			\label{fig: cau_grph_exmpl_pattern}
		\end{subfigure}
		\hfill
		\begin{subfigure}[b]{0.45\textwidth}
			\centering
			\includegraphics[width=\textwidth]{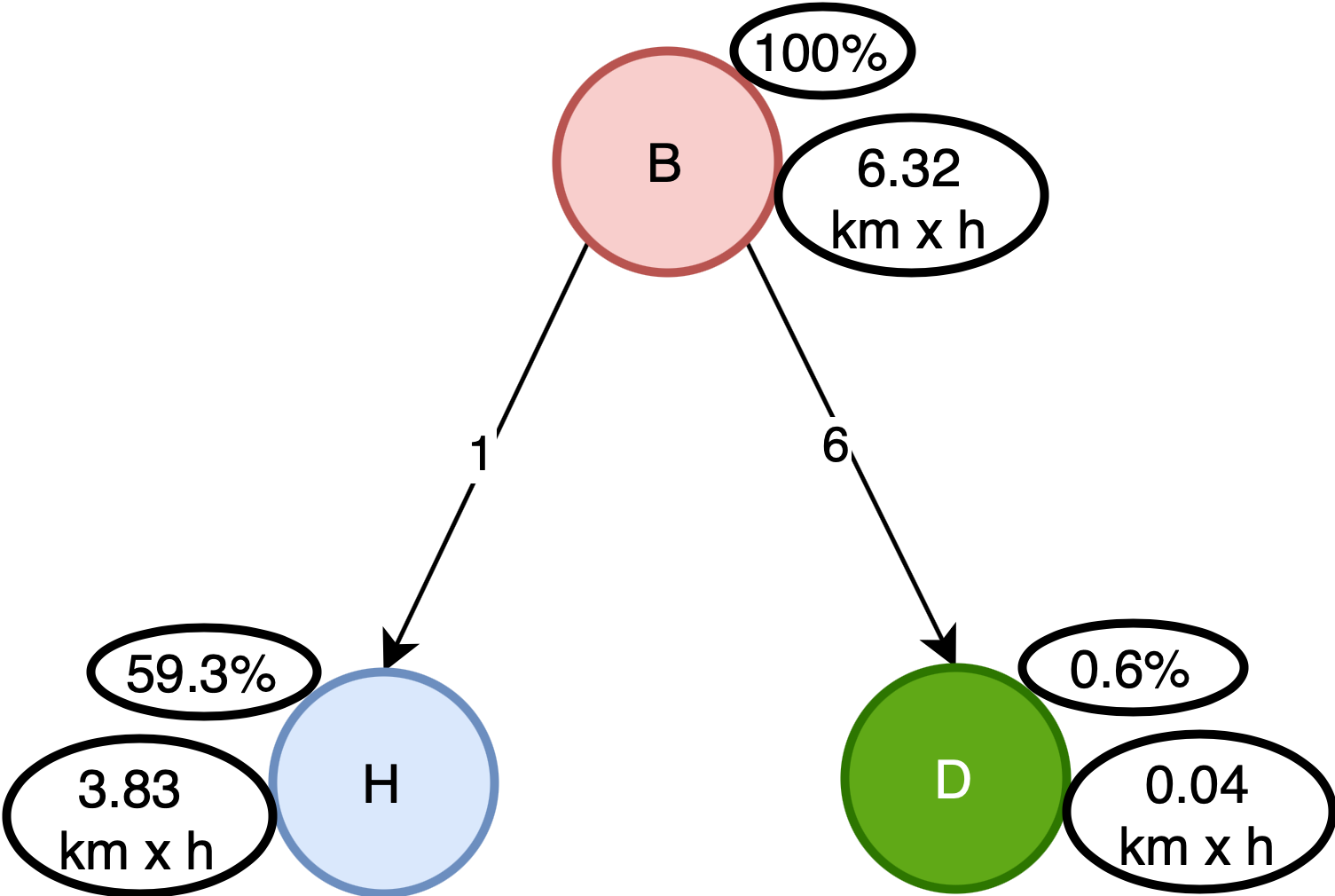}
			\label{fig: cau_grph_exmpl_grph}
		\end{subfigure}
		\caption{An example of relation graph: (a) a pattern of congested traffic at a bottleneck which causes heavily homogeneous congestion and later some small-scale disturbances, (b) its relation graph proposed by our method.}
		\label{fig:cau_grph_exmpl}
	\end{minipage}%
\end{figure}

Specifically, a node (or vertex) in our relation-graph represents a traffic phenomenon from a limited set of so-called \textit{traffic primitives}. We consider three such primitives and therefore three types of nodes in this study: these are the bottleneck (B), regions of homogeneous congestion (H) and disturbances (D), respectively. 
A node furthermore contains attributes describing the corresponding traffic primitive. In this study the main attribute of a node is \textit{size}, in either absolute form [km $\times$ hour] or relative form [proportion \%]. A directed link (or edge) represents a spatiotemporal relation between primitives (i.e. traffic components). This relation indicates a possible causality based on the observation that the starting point ${t,x}$ of one primitive is associated with the other primitive. For example, to represent many disturbances emerging from a single bottleneck, the corresponding relation graph has an edge from the related bottleneck node to the related disturbance node, with the edge weight indicating the number of disturbances. This representation results in compact relation-graphs (see Fig. \ref{fig:cau_grph_exmpl}).



A formal definition of this relation-graph is described in Definition \ref{def:relation_graph}.
\begin{definition}
	\label{def:relation_graph}
	The relation graph representing a congestion pattern is an attributed, directed graph $G=(E,V,A)$. Descriptions of these sets are as follows.
	\begin{equation*}
		\begin{split}
			\mathbf{V} &= \{v | v \text{ is a primitive}\} \\
			\mathbf{E} &= \{(v_i, v_j) | v_i \text{ (possibly) triggers } v_j\} \\
			\mathbf{A} &= (\tau, s^a, s^p, w) \text{ attribute set} \\
			&\tau: \mathbf{V} \rightarrow \{B - \text{bottleneck}, D - \text{disturbance}, H - \text{homogeneous congestion}\} \text{node label}\\
			&s^a:\mathbf{V} \rightarrow \mathbf{R} \text{ absolute size of a node }\\
			&s^p:\mathbf{V} \rightarrow \mathbf{R} \text{ relative size, i.e. the proportion (\%), of a node }\\
			&w:\mathbf{E} \rightarrow \mathbf{R} \text{ number of instances of the connection representing by the corresponding edge}
		\end{split}
	\end{equation*}
\end{definition}

Fig. \ref{fig:cau_grph_exmpl} illustrates the principle with the traffic pattern (left) and the resulting relation graph (right). The pattern shows traffic congestion at a bottleneck which is likely related to an incident. At the onset, traffic is heavily homogeneously congested. After some time, a few (minor) disturbances emerge before traffic regains free-flow conditions. The corresponding relation graph is constructed by identifying the three main elements in this pattern. These include the bottleneck - B node, homogeneous congestion node - H node, and disturbances - D node. Edges are associated with $w$. Specifically, the link (B-H) has a weight of 1 to represent 1 homogeneous region as shown in the pattern, whilst the link (B-D) has a weight of 6 that shows the number of disturbances detected. Furthermore, each node consists of attributes, namely absolute size and proportion.

\section{Similarity measurement}
\newcommand{\size}{a}
\newcommand{\freq}{w}

\label{sec_similaritymeasure}
The previous section shows how to construct an abstract and intuitive relation graph for a pattern of congestion. This section describes how similarities between congestion patterns are measured based on those graphs. Overall, (in-exact) graph matching is adopted, and a similarity function, which measures the resemblance between any pair of nodes in two respective graphs, is proposed. This function reflects several aspects of a pattern, including the similarities in the structure, and the proportions and frequencies of any extracted components. Details are given in the following paragraphs.

\subsection{Brief overview of graph matching}
By presenting congestion patterns using relation graphs, the measurement of similarities is transformed into graph similarity or so-called graph matching. There are two main categories in graph matching, namely exact graph matching and inexact graph matching (also known as error-tolerant graph matching) \citep{conte2004thirty,foggia2014graph,riesen2015structural,emmert2016fifty}. The former is strict in matching two graphs with respect to mapping their nodes or edges. This category is mainly intended for matching identical graphs. Meanwhile, the latter category is more flexible and allows differences in node/edge/subgraph mappings. In principle, these differences are tolerated with certain penalties when comparing graphs. This feature makes the second category more applicable to real-world applications where exact matching is not always guaranteed (if not rarely). In our application, matching relation graphs representing congestion patterns fall into the second category.

A graph matching is formulated as an optimisation problem, in which the cost of a matching function is generally defined as shown in Equation \ref{eq:general_matching_cost} (rewritten from \citep{foggia2014graph}). Note that, in inexact mapping, some nodes or edges of one graph might not have matches from the other graph. To formally describe this, a special so-called null node $\varepsilon$ is introduced. Accordingly, the mapping $f$ is annotated as $f: V_A \mapsto V_B \cup \{\varepsilon\}$. It is injective for nodes in $V_A$ that are not mapped to $\varepsilon$. For such nodes, the cost is called replacement cost $C^N_R$. Mapping a node to $\varepsilon$ is reasonably seen as the deletion of that node, and the related cost is called deletion cost $C^N_D$. Besides, edges are also needed to be mapped. Two similar types of mapping, i.e. replacement and deletion, are relevant to edge mapping and are evaluated by the cost functions $C^E_R, C^E_D$, respectively. Note that these individual cost functions are specialised, which means their definitions depend greatly on specific applications.

\begin{equation}
	\label{eq:general_matching_cost}
	\begin{aligned}
		C(f) = &\sum_{\substack{v \in V_A \\
				f(v) \ne \varepsilon}}C^N_R(v,f(v)) + 
		\sum_{\substack{v \in V_A \\
				f(v) = \varepsilon}}C^N_D(v) + 
		\sum_{\substack{v' \in V_B \\
				f^{-1}(v') = \varepsilon}}C^N_D(v') +\\
		&\sum_{\substack{e=(v_1,v_2)\in E_A \\
				e'=(f(v_1),f(v_2)) \in E_B}}C^E_R(e,e') + 
		\sum_{\substack{e=(v_1,v_2)\in E_A \\
				e'=(f(v_1),f(v_2)) \notin E_B}}C^E_D(e) +
		\sum_{\substack{e'=(v'_1,v'_2)\in E_B \\
				(f^{-1}(v'_1),f^{-1}(v'_2) \notin E_A}}C^E_D(e')
	\end{aligned}
\end{equation}

Various approaches have been proposed for graph matching by reformulating an optimisation problem on the cost function $C(f)$ such as graph edit distance \citep{bunke1997relation,gao2010survey}, graph kernels \citep{gartner2003graph}, iterative methods \citep{blondel2004measure,zager2008graph}. We refer to \citep{foggia2014graph,emmert2016fifty} for an in-depth survey of these approaches. Our work is motivated by the iterative approach. In principle, the similarities between nodes consider not only the two nodes but also their neighbour nodes. Hence, this approach, to some extent, combines the notations of different individual cost functions (Equation \ref{eq:general_matching_cost}) into one similarity function.

We propose a two-phase algorithm for measuring the similarity of two congestion patterns based on their relation graphs. Firstly, the similarities of all possible pairs of nodes between the two graphs are calculated. Secondly, the total similarity score of mapping all available nodes is optimised. The obtained score represents how similar the two patterns are. The following subsections describe these two terms in detail.

\subsection{Phase 1: Nodes similarity}
The similarity between two nodes (source nodes) is measured in a recursive way as motivated by \cite{zager2008graph}. Specifically, the similarity of subsequent nodes recursively contributes to the similarity score of their source nodes. Unlike in \citep{zager2008graph}, where scores from all possible pairs of nodes are accumulated, our proposed method only considers those from the best mapping between subsequent nodes.

\subsubsection{Similarity score for nodes}
The overall similarity score between two nodes, $n_A \in V_A, n_B \in V_B$ from $G_A,G_B$ respectively, is formulated as shown in Equation \ref{eq:node_distance}. The first part of the R.H.S., $S_0$, measures the similarity that is based intrinsically on their attributes (regardless of their neighbour nodes). The second part represents the accumulation of similarities from their subsequent nodes. 
Here, the parameter $\theta_i$ regulates how much of subsequent nodes' similarity attributes to the similarity of two source nodes. 
Note that the contribution of subsequent node similarities is, to some extent, equivalent to the similarity of matching the corresponding links (which is related to function $C^E_R$ in Equation \ref{eq:general_matching_cost}).

\begin{equation}
	\label{eq:node_distance}
	\begin{split}
		S(n_A, n_B) = S_0(n_A, n_B) + \theta_i \times min \big(S_0(n_A, n_B), &\argmax_{f:C_A \rightarrow C_B}\sum_{c_i^A \in C_A} S(c_i^A, f(c_i^A))\big)
	\end{split}
\end{equation}
where $C_A, C_B$ represents the sets of subsequent nodes of $n_A, n_B$, respectively.

Similar to the overall cost defined in Equation \ref{eq:general_matching_cost}, the base similarity $S_0$ captures several possibilities of node matching, which depend on whether both nodes are in the original graphs. Accordingly, two similar evaluations need to be defined, namely the so-called \textit{replacement} - $S_R(n_A, n_B)$ and \textit{deletion} - $C_D(n)$. Equation \ref{eq:base_sim_summary} summarises these cases.

\begin{equation}
	\label{eq:base_sim_summary}
	S_0(n_A, n_B) = \begin{cases}
		S_R(n_A, n_B), &\text{if $n_A \neq \varepsilon, n_B \neq \varepsilon$} \\
		-C_D(n_A), &\text{if $n_B = \varepsilon$} \\
		-C_D(n_B), &\text{if $n_A = \varepsilon$}
	\end{cases}
\end{equation}

There are two cases when matching two non-null nodes regarding whether they represent the same primitive type. If these nodes are different types, their mapping is equivalent to two deletion operations (see Equation \ref{eq:match_2_nonnull_nodes}).

\begin{equation}
	\label{eq:match_2_nonnull_nodes}
	S_R(n_A, n_B) = \begin{cases}
		M(n_A, n_B), &\text{if $\tau(n_A) = \tau(n_B)$} \\
		-C_D(n_A)-C_D(n_B), &\text{if $\tau(n_A) \neq \tau(n_B)$}
	\end{cases}
\end{equation}

The previous setup leads to defining two basic functions, i.e. $M(n_A, n_B)$ and $C_D(n)$. Choices for these functions are specialised with respect to specific applications. In our proposed framework, we formulate these functions with respect to the selected attributes associated with nodes/edges in relation graphs. Also, these functions are parameterised by utilising certain parameters. The objective is to inject different perspectives when looking for similar characteristics from congestion patterns.

\subsubsection{Balancing similarity and differences}
Our proposed function for measuring similarity between two commonly labelled nodes accounts for both the resemblance between their attributes and the importance of their difference. For that, the designed function includes both their overlapping size and the size of the referenced node. The former acts as a proxy to the similarity of the two nodes. The latter is used to compensate for the difference (if any) between the two nodes. The first node is selected as a referenced node in our set-up. The parameter $\theta_g$ regulates the scales of these two terms (see Equation \ref{eq_award_similarity}).

The detailed similarity between two nodes is measured based on the overlapping size. Note that a different function is possible when different properties are used for node attributes. On the other hand, the unmatched size is also taken into account as this assists in ranking the closeness of different pairs of nodes. In particular, a logistic function is formulated to translate the size difference (in terms of proportions to the total size) to a number (i.e. weight) that scales the overall similarity. The contribution of this difference is regulated by the parameter $\theta_s \ge 0$ (see Equation \ref{eq_award_similarity}). In addition, the difference in the occurrences ($w$) of the two nodes is also dealt with. A 'virtual node' $n_E$, with relevant features, $a$ (see Equation \ref{eq:size}) and $w$, is created as shown in Equation \ref{eq:node_w_difference}. The underlying idea is to apply deletion cost $C_D(n_E)$ to the occurrence difference when matching two nodes. Parameter $\theta_w \ge 0$ regulates the tolerance of this difference.


\begin{equation}
	\begin{split}
		M(n_A,n_B) = &(1-\theta_g) * \big[ 2 \times w_{min}\times \size_{min} \times \mathcal{L}_{\beta_1, \beta_0}(\theta_s, \frac{\Delta a}{\sum a}) - C_D(n_E)\big] \\
		& + \theta_g \times 2 \times w(n_A) \times a(n_A)
	\end{split}
	\label{eq_award_similarity}
\end{equation}

where,

\begin{alignat}{3}
	&\text{Common size }&&\size_{min} &&= min \big( \size(n_A), \size(n_B) \big)\\
	&\text{Size difference }&&\Delta a &&= |\size(n_A) - \size(n_B)| \label{eq:delta_size}\\
	&\text{Total size }&&\sum a &&= \size(n_A) + \size(n_B) \label{eq:total_size}\\
	&\text{Logistic function }&&\mathcal{L}_{\beta_1, \beta_0}(\theta, x) &&= 1 - \frac{1}{1 + e^{\beta_1(\theta x) + \beta_0}} \label{eq:logistic} \\
	&\text{Occurrence-difference node }&&n_E &&\begin{cases}
		\size = \begin{cases}
			\size(n_A) \text{, if } \freq(n_A) > \freq (n_B) \\
			\size(n_B) \text{, if otherwise}
		\end{cases} \\
		\freq = f_{\beta_1,\beta_0}(\theta_w, \Delta \freq)
	\end{cases} \label{eq:node_w_difference} \\
	&\text{Occurrence difference }&&\Delta \freq &&= |\freq(n_A) - \freq(n_B)| \label{eq:delta_weight} \\
	&\text{Size selection: }&&\size(n) &&= \begin{cases}
		s^a(n), &\text{ for absolute size, i.e. area (km$\times$minute)} \\
		s^p(n), &\text{ for proportion (\%)} 
	\end{cases}\label{eq:size}
\end{alignat}


\subsubsection{Node deletion cost}
As overlapping sizes are used for attributing commonly labelled nodes, the cost of deleting a node can be justified by its size. A parameter $\theta_d$ is introduced here to regulate how much penalty is applied for not finding a match for a node. Equation \ref{eq_penalise_dissimilarity} gives a definition of this cost.

\begin{equation}
	\label{eq_penalise_dissimilarity}
	C_D(n) = \theta_{t} \times \size(n)
\end{equation}

%

Table \ref{tbl:params_node_dist} summarises all the parameters and their meanings in customising a similarity measurement between any pair of nodes.

\begin{table}[h]
	\centering
	\caption{Parameters for customising similarity measurement between relation-graphs}
	\label{tbl:params_node_dist}
	\begin{tabular}[t] {p{2cm} p{11cm}}
		\hline
		\textbf{Parameter} & \textbf{Description} \\
		\hline
		$\theta_s$ & Penalise size difference \\
		$\theta_g$ & Regulates the trade-off between node size match (maximised when $\theta_g = 0$) and node type match (maximised when $\theta_g = 1$) \\
		$\theta_d$ & Penalise node type difference, therefore, regulate the tolerance of having unmatched nodes \\
		$\theta_w$ & Penalise the differences in frequency attribute: whether to focus on overall structure or details \\
		$\theta_i$ & Regulate the contribution of subsequent-node similarities to the matching of two source nodes \\
		\hline
	\end{tabular}
\end{table}

\subsection{Phase 2: Nodes mapping}

Given two relation graphs that represent two congestion patterns, the previous section shows how to measure the similarity between any pairs of nodes therein. This section describes how to come up with a similarity score at the pattern level.

\begin{figure}
	\centering
	\begin{minipage}{0.8\textwidth}%
		\centering
		\qquad
		\includegraphics[width=0.7\textwidth]{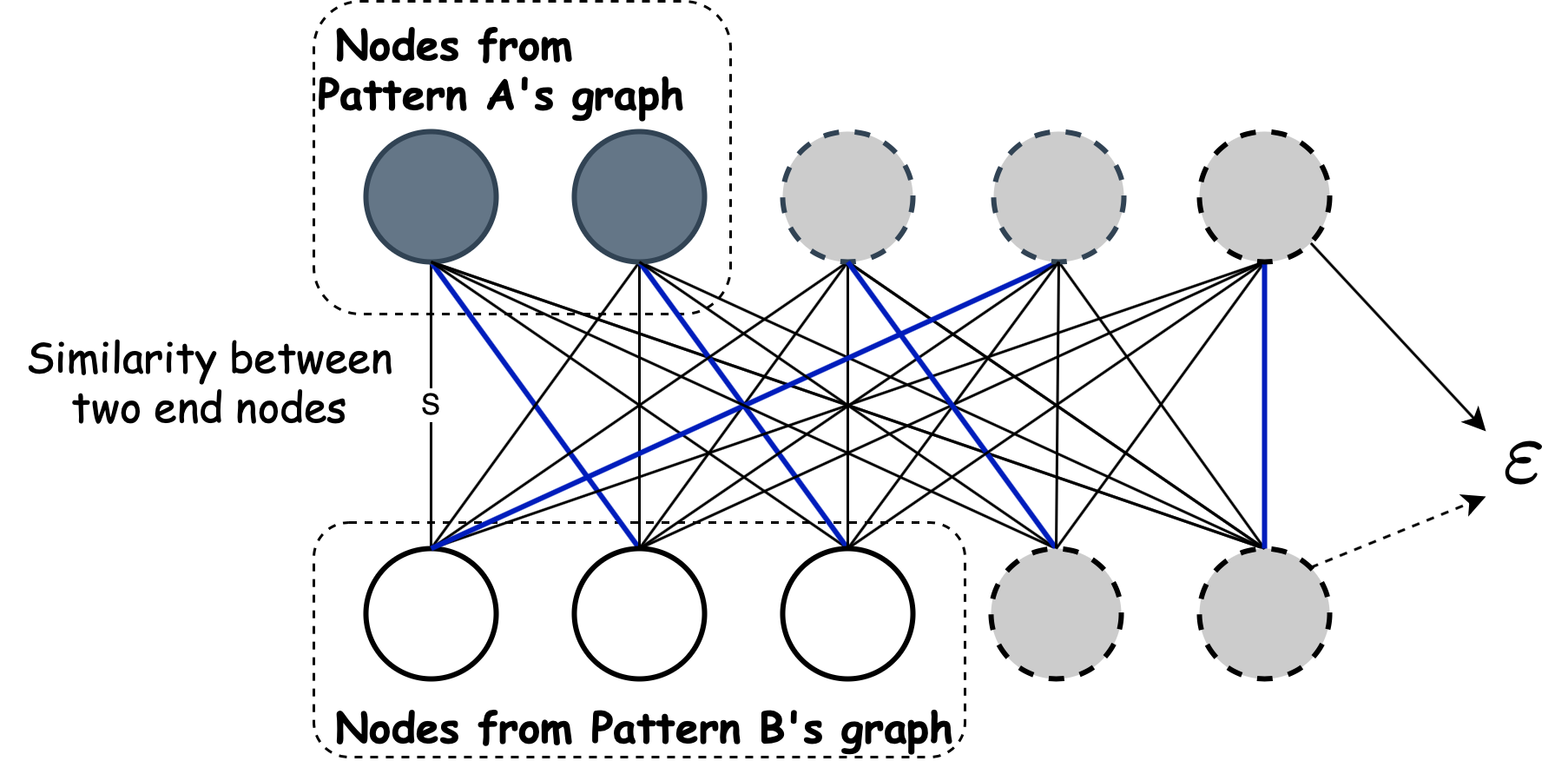}
		\caption{An illustration of how to formulate the pattern matching as an assignment problem between their node sets. Edges' weights are the similarities between the corresponding end nodes using Equation \ref{eq:node_distance}. A feasible assignment is highlighted in blue colour, in which one node is exactly matched to another node.}
		\label{fig:HungarianAlgIllus}
	\end{minipage}%
\end{figure}

To evaluate how the two patterns match, we formulate the problem as an assignment problem which finds the so-called perfect matching between nodes from the two graphs. That assignment maximises the total scores from all pairs of matched nodes under the condition that one node is matched with exactly another one. This perfect matching (once found) is considered the best mapping between the two source nodes. The corresponding total score then indicates the similarity between the two patterns.  An illustration of our assignment problem is depicted in Fig. \ref{fig:HungarianAlgIllus}. A complete bipartite graph is constructed to show all possible mapping of nodes from two graphs. The weight of each edge is associated with the similarity score of corresponding nodes. The solution of pattern mapping is the perfect matching with the maximum total edge' weights. Equation \ref{eq:assignment_prob} formulates this assignment approach in mathematical terms.

\begin{equation}
	\label{eq:assignment_prob}
	\begin{aligned}
		S(p_A, p_B) &= \argmax_{f:\Omega_A \rightarrow \Omega_B} \sum_{n \in \Omega_A} S(n, f(n)) \\
		\text{where,} \\
		\Omega_A &= V_A \cup \{\varepsilon,...,\varepsilon\} \\
		\Omega_B &= V_B \cup \{\varepsilon,...,\varepsilon\} \\
		|\Omega_A| &= |\Omega_B| = |V_A| + |V_B|
	\end{aligned}
\end{equation}

The assignment problem is solved by applying the well-known Hungarian algorithm (also known as the Kuhn-Munkes algorithm), which was developed by \cite{kuhn1955hungarian}. It has polynomial complexity, in particular, $\mathcal{O}(n^3)$.

\section{Experiment results and discussion}
In this section, we demonstrate the ability of the proposed retrieval framework to retrieve similar patterns from a collection of traffic congestion patterns. The analysis includes three aspects. First, some exemplary queries are conducted, and their performances are investigated with respect to the corresponding obtained patterns. Second, we discuss the impacts of tuning the parameters (in Table \ref{tbl:params_node_dist}) for reflecting different perspectives on similarities between patterns. Third, we consider the computational complexity as well as its implication in applying to large datasets of the proposed method. Details are in the following sections.

\label{sec_experimentandresult}
\subsection{Data \& parameter settings}
\begin{figure}
	\centering
	\begin{minipage}{0.8\textwidth}%
		\centering
		\includegraphics[width=0.7\textwidth]{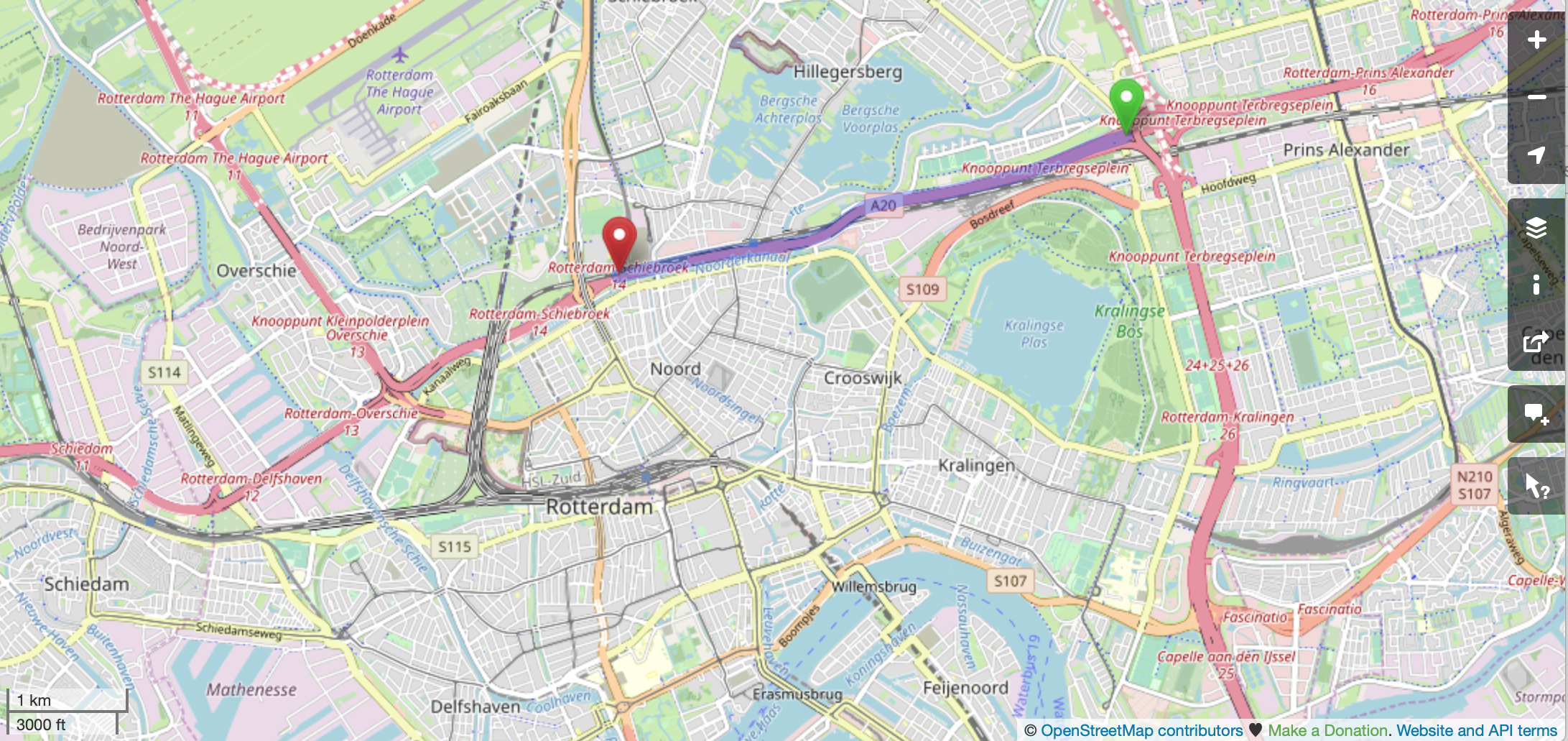}
		\caption{A broad view of the selected corridor in the experiment. The image is taken from Open Street Map (and reproduced from \citep{nguyen2021automated})}.
		\label{fig:a16a20map}
	\end{minipage}%
\end{figure}

To evaluate the proposed method, we have selected a corridor on the ring of Rotterdam, which is one of the busiest roadways in the Netherlands. Fig. \ref{fig:a16a20map} shows a broad view of the road. It is approximately four kilometres long and comprises several active bottlenecks. These bottlenecks and downstream bottlenecks have caused much recurrent traffic congestion, therefore, it is a suitable choice for evaluating our proposed framework.

Speed data are provided by the National Data Warehouse (NDW), the Netherlands \cite{ndw}, in which each measurement is a one-minute aggregation of speed surpassing the related induction-loop detector's implemented location. To have a better view of resulting traffic, we apply the ASM method (Adaptive Smoothing Method) \citep{treiber2002reconstructing, schreiter2010two} to estimate speeds at finer resolutions both spatially and temporally, namely 100 meters by 30 seconds. We have processed data from the entire year of 2018 to obtain 778 patterns, which constitute the collection of traffic congestion patterns for evaluating our proposed method.

For the similarity measurement, the settings for all parameters are given in Table \ref{tbl:params_settings}. By setting $\theta_t, \theta_s, \text{and } \theta_w$ to 1, the total differences in type, size, and frequency, respectively, are fed to the logistic function to measure related penalties. As $\theta_i$ is set to 1, similarities from subsequent nodes are accumulated to the corresponding president nodes. This, to some extent, takes pattern structure into consideration. Therefore, we set $\theta_g$ to 0 for simplifying the base similarity function. This leads to a full assessment of related attributes when matching two nodes. Chosen values of $\beta$ set the changing point of the corresponding logistic function at the middle of input ranges. Note that there are no strict regulations in selecting these parameters. The presented settings are one of many possibilities and have led to good results in our experiment.

\begin{table}[h]
	\centering
	\caption{Parameter settings in the conducted experiment}
	\label{tbl:params_settings}
	\begin{tabular}[t] {c||c}
		\hline
		\textbf{Parameter} & \textbf{Value} \\
		\hline
		$\theta_s$ &  1\\
		$\theta_g$ &  0\\
		$\theta_t$ & 1 \\
		$\theta_w$ & 1 \\
		$\theta_i$ & 1 \\
		\hline
		\multicolumn{2}{c}{\textit{logistics} $\mathcal{L}$} \\
		$(\beta_1, \beta_0)$ & (10, -5) \\
		\hline
	\end{tabular}
\end{table}

\subsection{Retrieval results}
To demonstrate the feasibility of the proposed relation graph in the retrieval application, we analyse some example queries, namely for single disturbance, stop-and-go congestion, homogeneous congestion and a mix of these. These are typical patterns of congestion that are commonly observed in traffic data \citep{helbing2009theoretical, nguyen2016traffic, krishnakumari2017traffic, nguyen2019feature}. Finding their occurrences provides meaningful information for various purposes. For instance, to analyse how repetitively these types of congestion occur, and possible variations therein. This leads to a more thorough understanding of popular types of congestion. Another application for this is traffic model development, specifically model evaluation. Different scenarios, which are derived from similar patterns, can be tested with respect to the occurrences of certain congestion patterns.

\subsubsection{Single disturbance retrieval}
\begin{figure}
	\centering
	\begin{minipage}{0.8\textwidth}%
		\centering
		\includegraphics[width=1\textwidth]{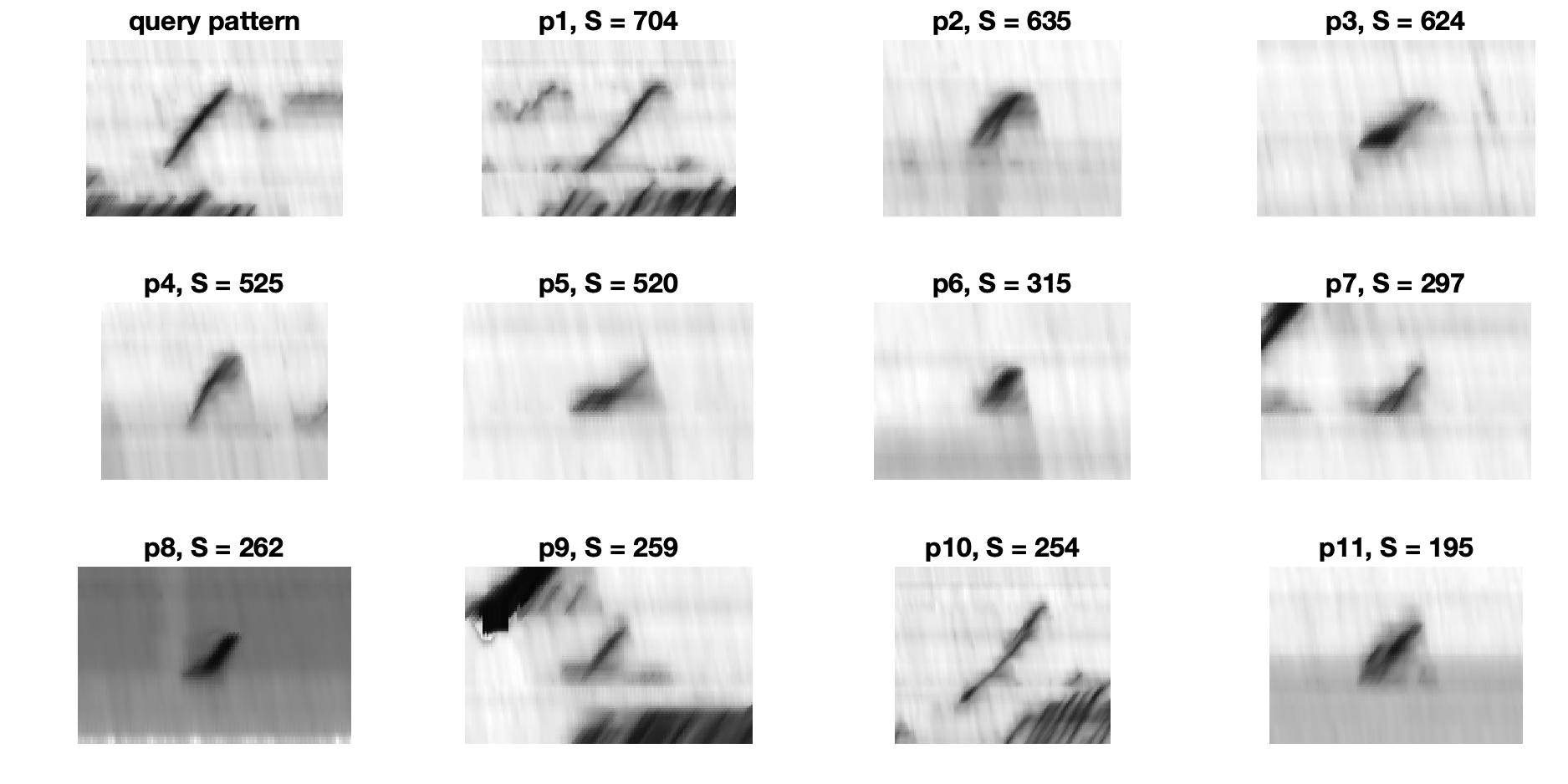}
		\caption{The 11 most similar patterns returned from searching for a moving disturbance (shown in the top-left pattern). Patterns are shown in the same resolution, hence, their size differences can be relatively shown. Note that, regions of congestion at the edges of some patterns should be ignored because they are the results of cropping out the patterns, i.e. they are not included as (main) content of the patterns. Besides, similarity scores are given as $S$ for each of the patterns.}
		\label{fig:retrieve_disturb}
	\end{minipage}%
\end{figure}

Figure \ref{fig:retrieve_disturb} shows an example of retrieving patterns representing a single disturbance. The implemented framework successfully returned patterns representing small disturbances as indicated in the query pattern. Note that, as disturbances are one of the nodes in the relation graph, this retrieval outcome is a direct result of the extracting method. Regarding the order of these patterns, some might seem more similar than those in higher ranks. For example, the pattern p4 seems more resembling the query pattern than the above two patterns (in the order list). The reason is that by choosing areas as an attribute, we have reduced two dimensions, i.e. spatial and temporal, down to only one. Therefore, introducing propagating lengths as attributes for disturbance nodes could fine-tune the results further. Nonetheless, in this work, we use the same attributes for all the nodes to simplify the graph and focus on demonstrating the feasibility of the proposed approach. We leave this enhancement for future work.

\newcommand{\expdir}{figs/exp/v5}

\subsubsection{Stop-and-go congestion retrieval}
\begin{figure}
	\centering
	\begin{minipage}{0.8\textwidth}%
		\centering
		\includegraphics[width=1\textwidth]{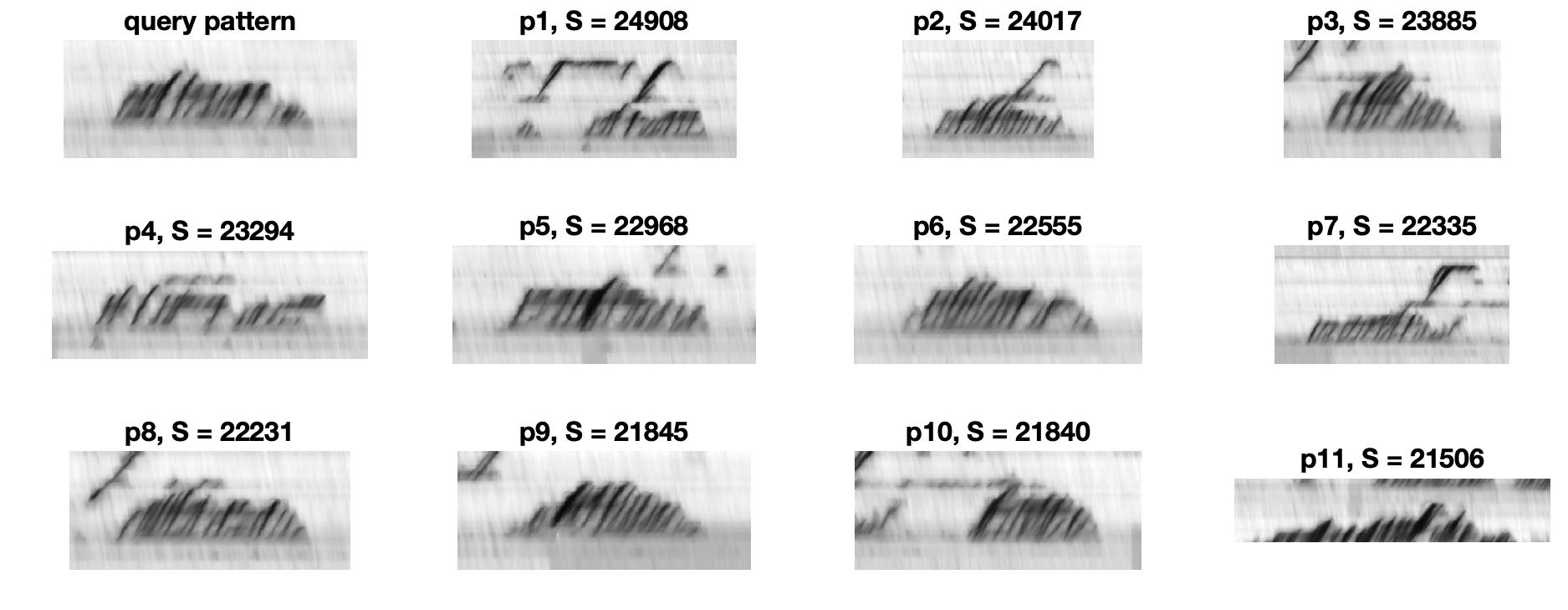}
		\caption{Retrieval results of stop-and-go congestion.}
		\label{fig:retrieve_oscil}
	\end{minipage}%
\end{figure}

Stop-and-go traffic waves is another common type of congestion where multiple disturbances occur over time. An example of retrieving such patterns is shown in Fig. \ref{fig:retrieve_oscil}. In the query pattern, a bottleneck is activated, from which many disturbances emerge. All the obtained patterns represent the same traffic phenomena. By detecting both the primary bottlenecks and probably the minor upstream secondary bottleneck, along with multiple disturbances, the obtained relation graphs are effective for locating patterns with the same topology as in the query.

\subsubsection{Homogeneous congestion retrieval}
\begin{figure}
	\centering
	\begin{minipage}{0.8\textwidth}%
		\centering
		\includegraphics[width=1\textwidth]{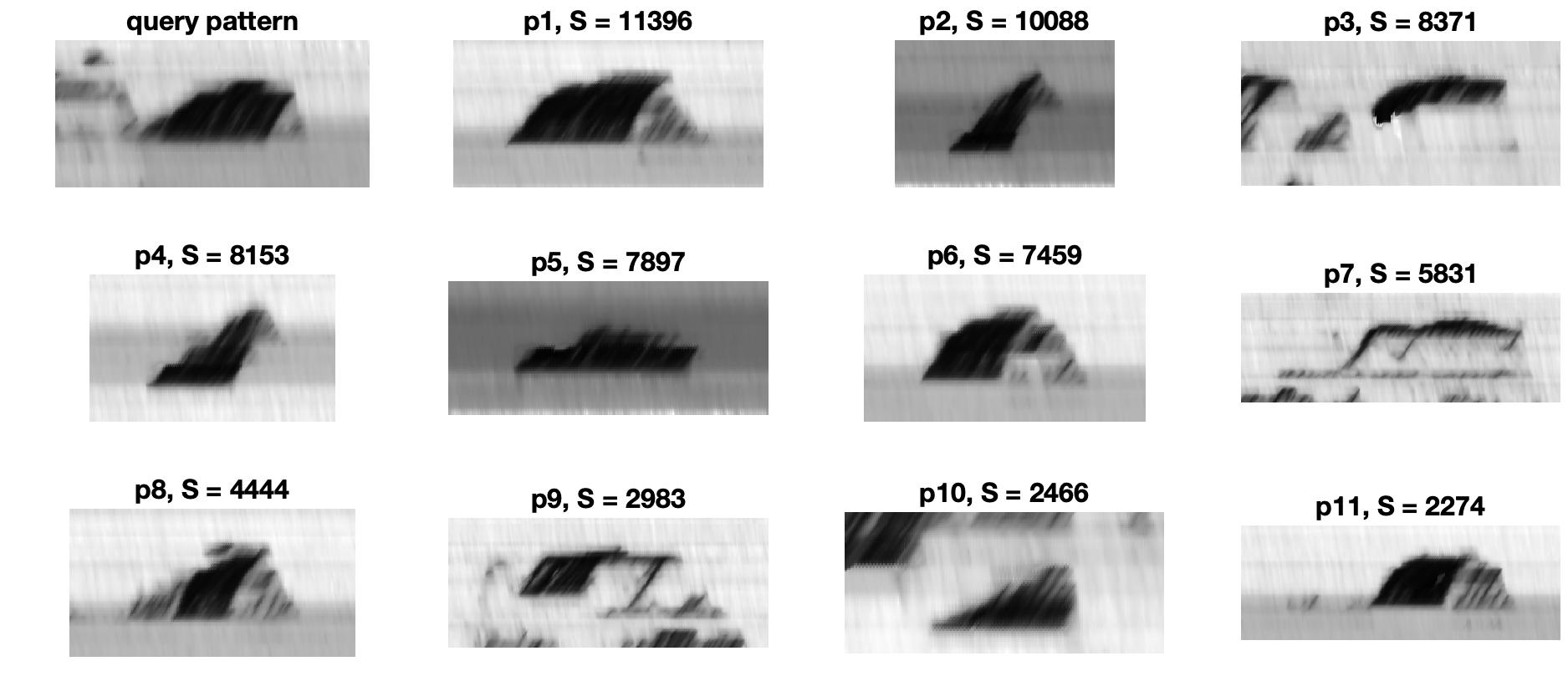}
		\caption{Retrieval results of homogeneous congestion.}
		\label{fig:retrieve_homo}
	\end{minipage}%
\end{figure}

An example of retrieving homogeneous congestion is illustrated in Fig. \ref{fig:retrieve_homo}. The given pattern represents significantly slow traffic upstream of a bottleneck (probably due to incidents like accidents). Hence, the two most important components of the corresponding relation graph for this pattern are a bottleneck node and a homogeneity node. Overall, the obtained patterns do represent the main phenomenon. This effective retrieval is a direct outcome of the extracting method for domain-specific features.

Note that the shapes of homogeneous areas in the obtained patterns are not necessarily identical to that of the ones in the query pattern. This is explained by the fact that the currently chosen attribute includes only sizes. 

\subsubsection{Complex congested traffic retrieval}
\begin{figure}
	\centering
	\begin{minipage}{0.8\textwidth}%
		\centering
		\includegraphics[width=1\textwidth]{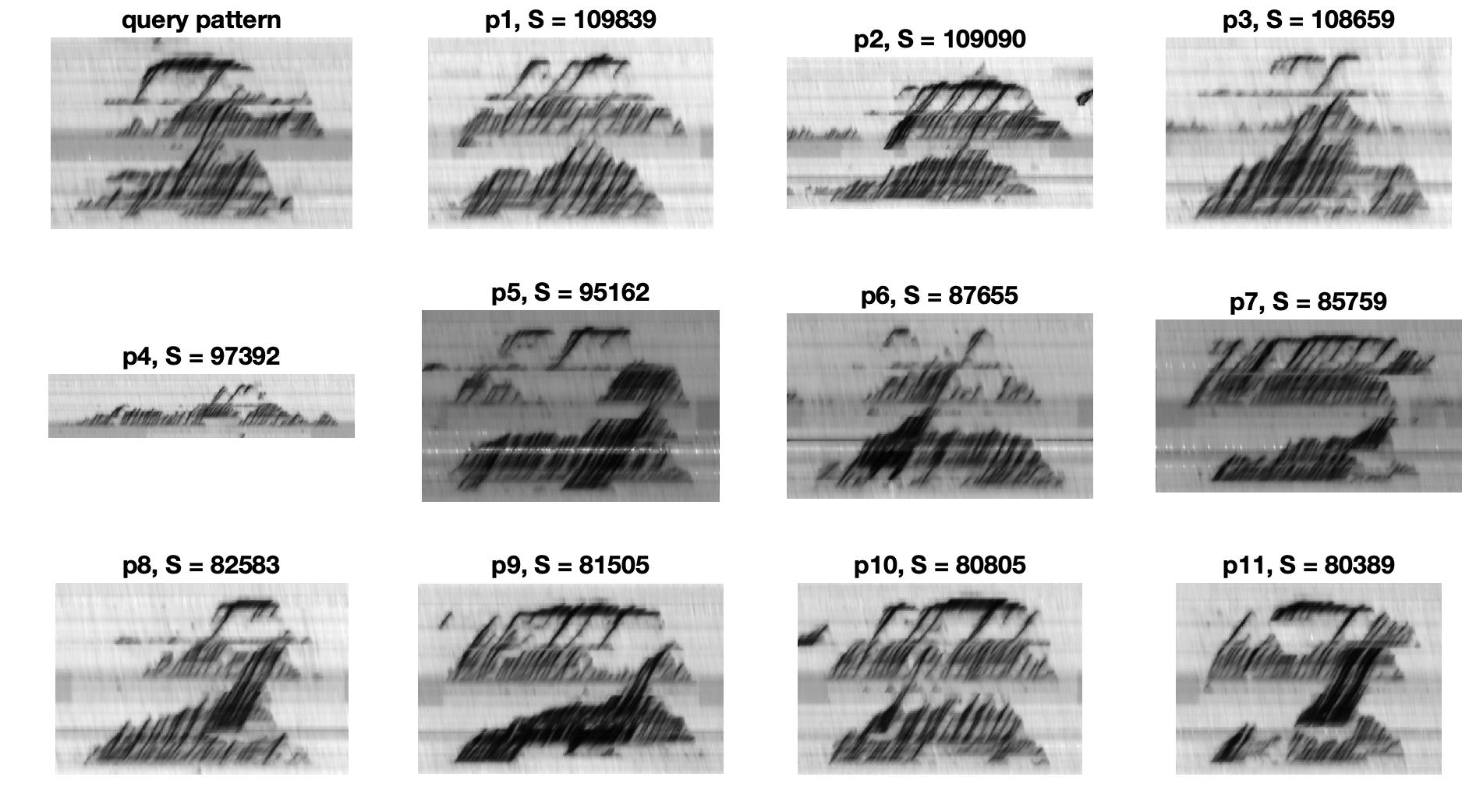}
		\caption{Retrieval results of \textit{meta} congestion.}
		\label{fig:retrieve_complex}
	\end{minipage}%
\end{figure}

Fig \ref{fig:retrieve_complex} illustrates an attempt to retrieve large-scale congestion patterns. The query pattern consists of various types of traffic jams, including disturbances that occur fairly frequently, multiple bottleneck activations, and a homogeneous congested area. Many obtained patterns can cope with these complications in the input pattern, meaning they have different activations of bottlenecks that cause dense stop-and-go traffic. Some of them show homogeneous regions. Regarding the overall structure, several patterns (for instance, p1, p2 or p3) represent two clusters of disturbances that are potentially due to the activations of two primary bottlenecks. 

\subsection{Parameter impacts}
In this section, we analyse how modifying parameters can change similarity scores and, hence, alter the ranks of obtained patterns. This is relevant for revising retrieval results in case a result is not as expected. The list of all parameters is shown in Table. \ref{tbl:params_node_dist}.
\subsubsection{Size penalty $\theta_s$}
\begin{figure}
	\centering
	\begin{minipage}{0.8\textwidth}%
		\centering
		\includegraphics[width=1\textwidth]{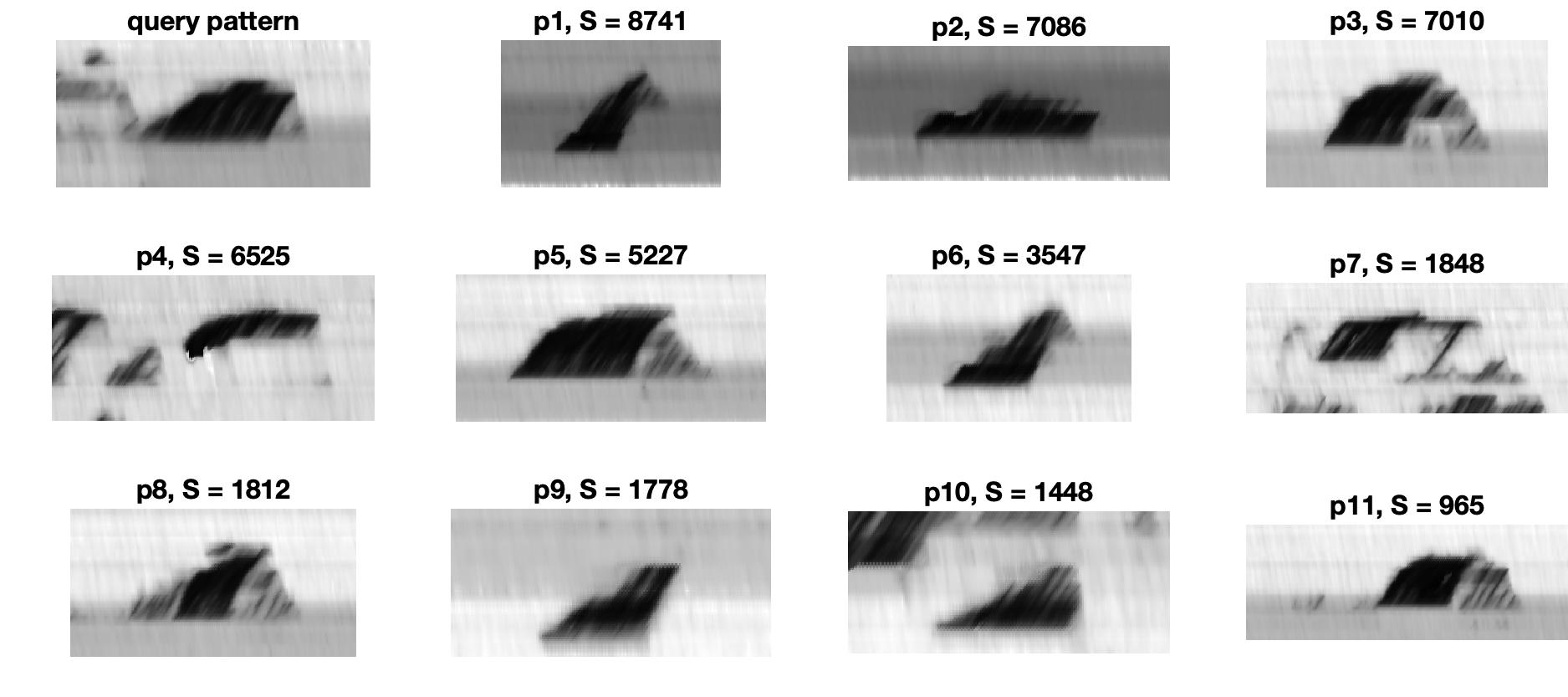}
		\caption{Another retrieval result of the homogeneous congestion in Fig. \ref{fig:retrieve_disturb}) with different parameter $\theta_s = 2$.}
		\label{fig:param_s_retrieve_homo}
	\end{minipage}
\end{figure}

The parameter $\theta_s$ penalises the difference in sizes between two matched nodes. Therefore, it regulates how important is to search for nodes of similar types. To demonstrate this, we modify $\theta_s = 2$ for the retrieval in Fig. \ref{fig:retrieve_homo}. This modification enforces a stricter condition on the sizes of matching nodes. The corresponding result is shown in Fig. \ref{fig:param_s_retrieve_homo}. Overall, the similarity scores of returned patterns decrease. The order of patterns consists of various changes such as the promoting of p1 (from the 2\textsuperscript{nd} to the 1\textsuperscript{st} place) and p5 (from the 1\textsuperscript{st} to the 5\textsuperscript{th}). In addition, new patterns are also moved forward, such as p10. 

\subsubsection{Weight penalty}
The parameter $\theta_w$ controls the frequency of a component's appearances. This is mostly relevant to disturbances in stop-and-go traffic patterns. By increasing or decreasing this parameter, the outcomes are adjusted to be against or in favour of the differences in the frequencies of disturbances. 

\begin{figure}
	\centering
	\begin{minipage}{0.8\textwidth}%
		\centering
		\includegraphics[width=1\textwidth]{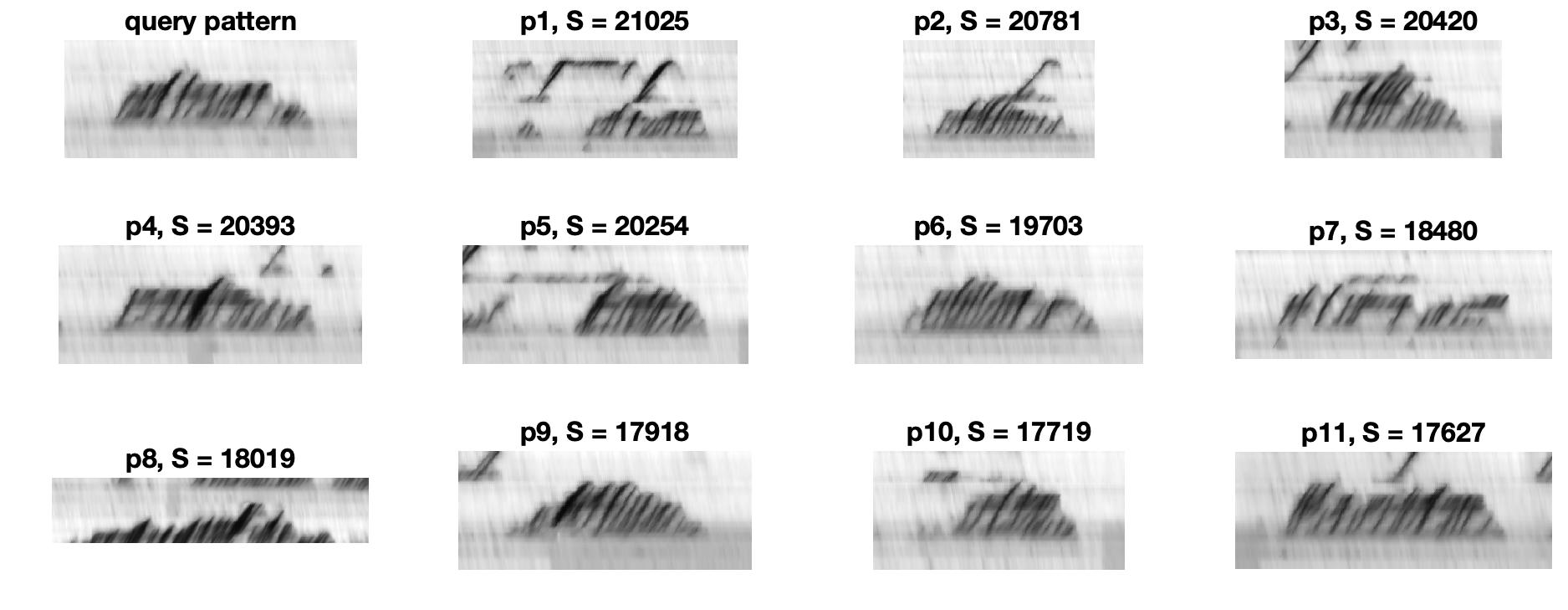}
		\caption{Another retrieval result of the stop-and-go congestion in Fig. \ref{fig:retrieve_oscil} by increasing $\theta_w$ to 3.}
		\label{fig:param_w_retrieve_oscil}
	\end{minipage}
\end{figure}

Fig. \ref{fig:param_w_retrieve_oscil} demonstrates the impact of increasing $\theta_w$ on the search made in Fig. \ref{fig:retrieve_oscil}. Even though there is not much (overall) difference compared to the previous result, this new result shows several changes in the order.  The new ranking promotes those patterns with more similar numbers of disturbances as in the example pattern. The overall similarity scores are smaller due to the stricter condition of occurrence frequencies.

\subsubsection{Unmatch penalty}
There may be unmatched nodes from two relation graphs. How much this decreases the similarity score is regulated by the parameter $\theta_d$. By lowering this parameter, users opt for finding the completion of the components in the query pattern, and at the same time tolerate the existence of extra components in the target patterns. Similarly, increasing $\theta_d$ aims for the compact of target patterns with respect to the given pattern.

\begin{figure}
	\centering
	\begin{minipage}{0.8\textwidth}%
		\centering
		\includegraphics[width=1\textwidth]{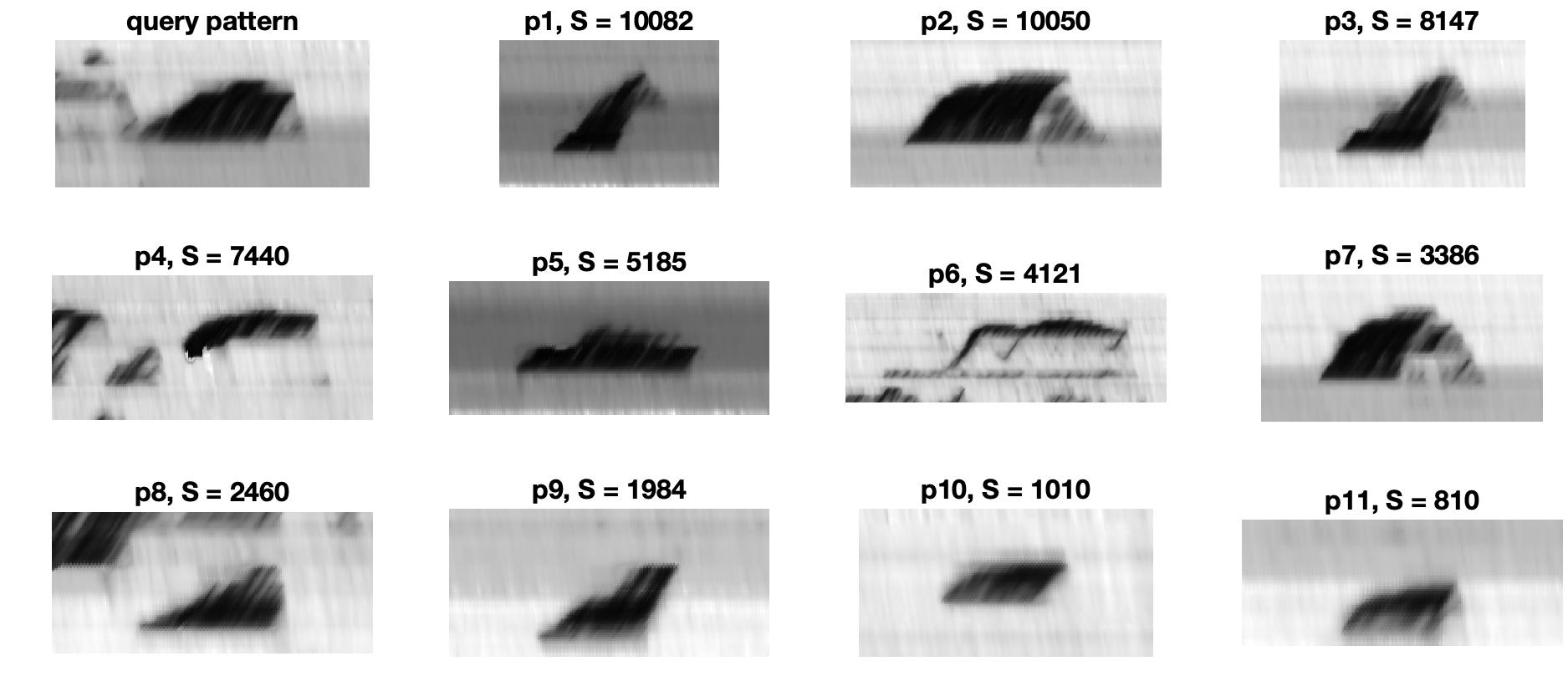}
		\caption{Another retrieval result of the homogeneous congestion in Fig. \ref{fig:retrieve_homo} by increasing $\theta_d$ to 2.}
		\label{fig:param_t_retrieve_homo}
	\end{minipage}
\end{figure}

An example of the effect of increasing $\theta_d$ is shown in Fig. \ref{fig:param_t_retrieve_homo}, which is a modified retrieval of the one in Fig. \ref{fig:retrieve_homo}. Since $\theta_d$ has a higher value, those patterns with a more compact representation of homogeneous regions, less other extra regions, are advanced in the ranking list.

\subsubsection{Structural integrity}
\begin{figure}
	\centering
	\begin{minipage}{0.8\textwidth}
		\centering
		\begin{subfigure}[h]{\textwidth}
			\centering
			\includegraphics[width=\textwidth]{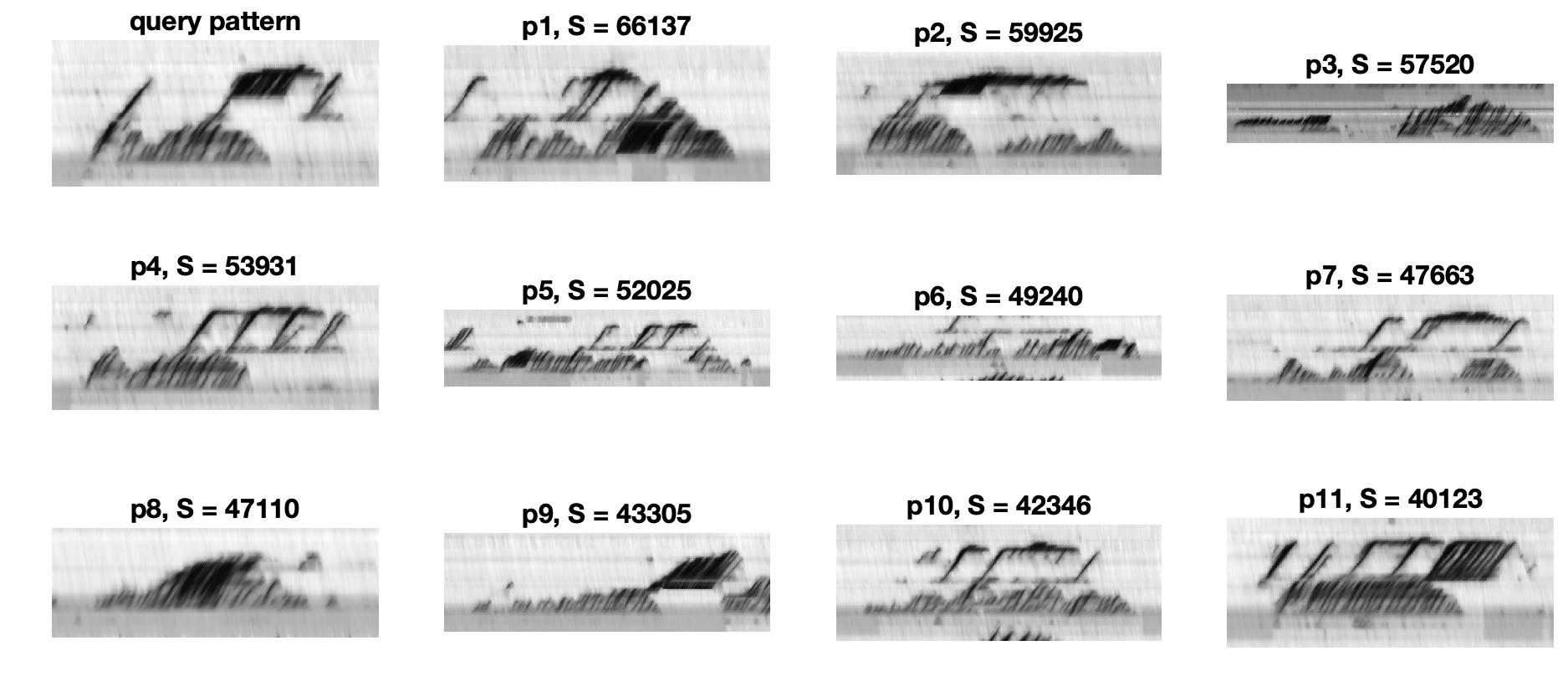}
			\caption{$(\theta_s, \theta_g \theta_d, \theta_w, \theta_i) = (1,0,1,1,1)$}
			\label{fig:param_I_retrieve_complex_good:noI}
		\end{subfigure}
		\begin{subfigure}[h]{\textwidth}
			\centering
			\includegraphics[width=\textwidth]{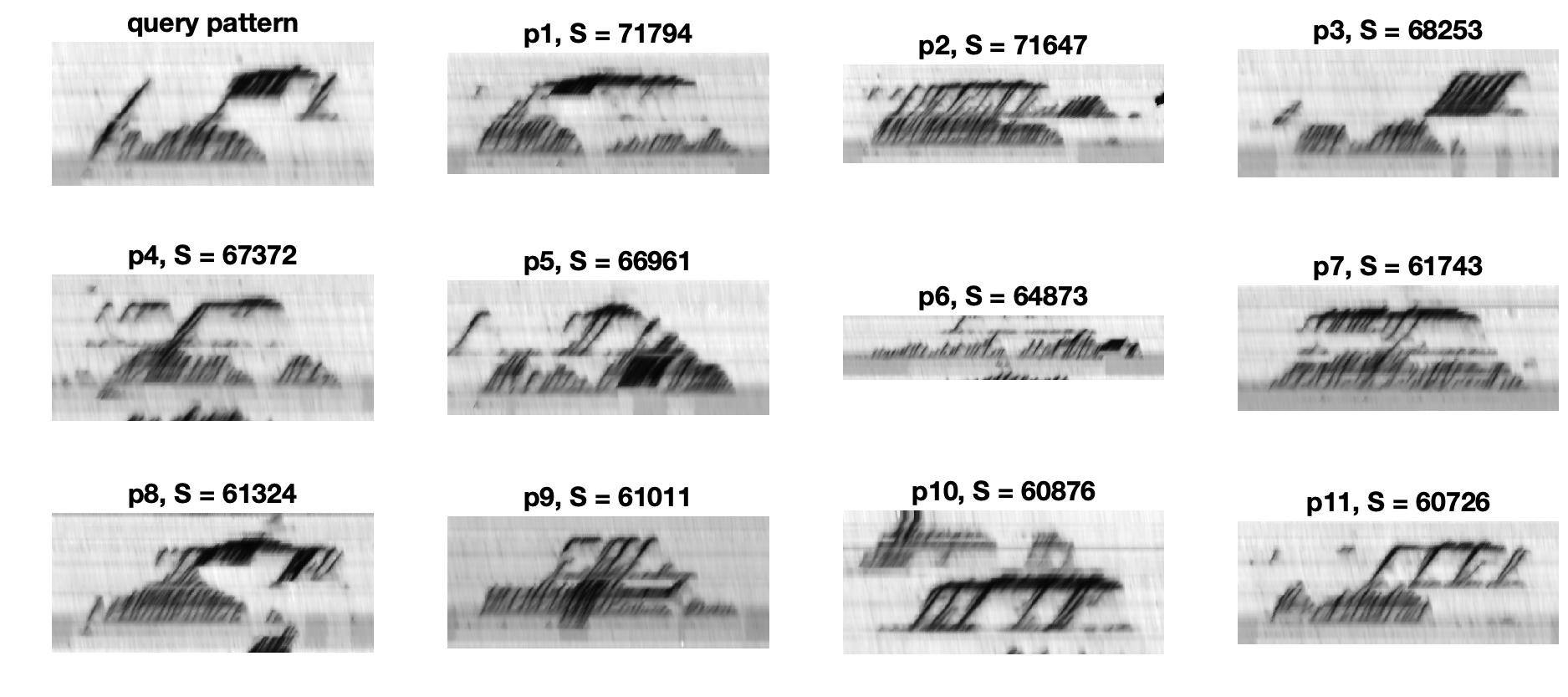}
			\caption{$(\theta_s, \theta_g \theta_d, \theta_w, \theta_i) = (1,0.7,1,1,1)$}
			\label{fig:param_I_retrieve_complex_good:yesI}
		\end{subfigure}
		\caption{The effect of the parameter $\theta_i$.}
		\label{fig:param_I_retrieve_complex_good}
	\end{minipage}
\end{figure}

The parameters $\theta_g, \theta_i$ are designed to promote the matching of pattern structures. A demonstration of their use is illustrated in Fig. \ref{fig:param_I_retrieve_complex_good}. Fig. \ref{fig:param_I_retrieve_complex_good:noI} shows the example in which both similarities of pairs of matched nodes and their subsequent nodes are relatively important. 
On the other hand, by setting $\theta_g = 0.7$, the importance of having the same structure becomes higher while that of node similarities is reduced. The obtained patterns in Fig. \ref{fig:param_I_retrieve_complex_good:yesI} demonstrate the effect of this change. Differences between components of the obtained patterns and those in the example patterns are more tolerant. As a result, some good similar patterns are advanced to the top list, e.g. p1, p3, p4, p8. Note that, increasing $\theta_i$ leads to low importance levels of node features. This, therefore, can result in patterns that are quite different from the query example despite sharing a common structure.

\subsection{Time complexity}
\begin{figure}
	\centering
	\begin{minipage}{0.7\textwidth}%
		\centering
		\includegraphics[width=1\textwidth]{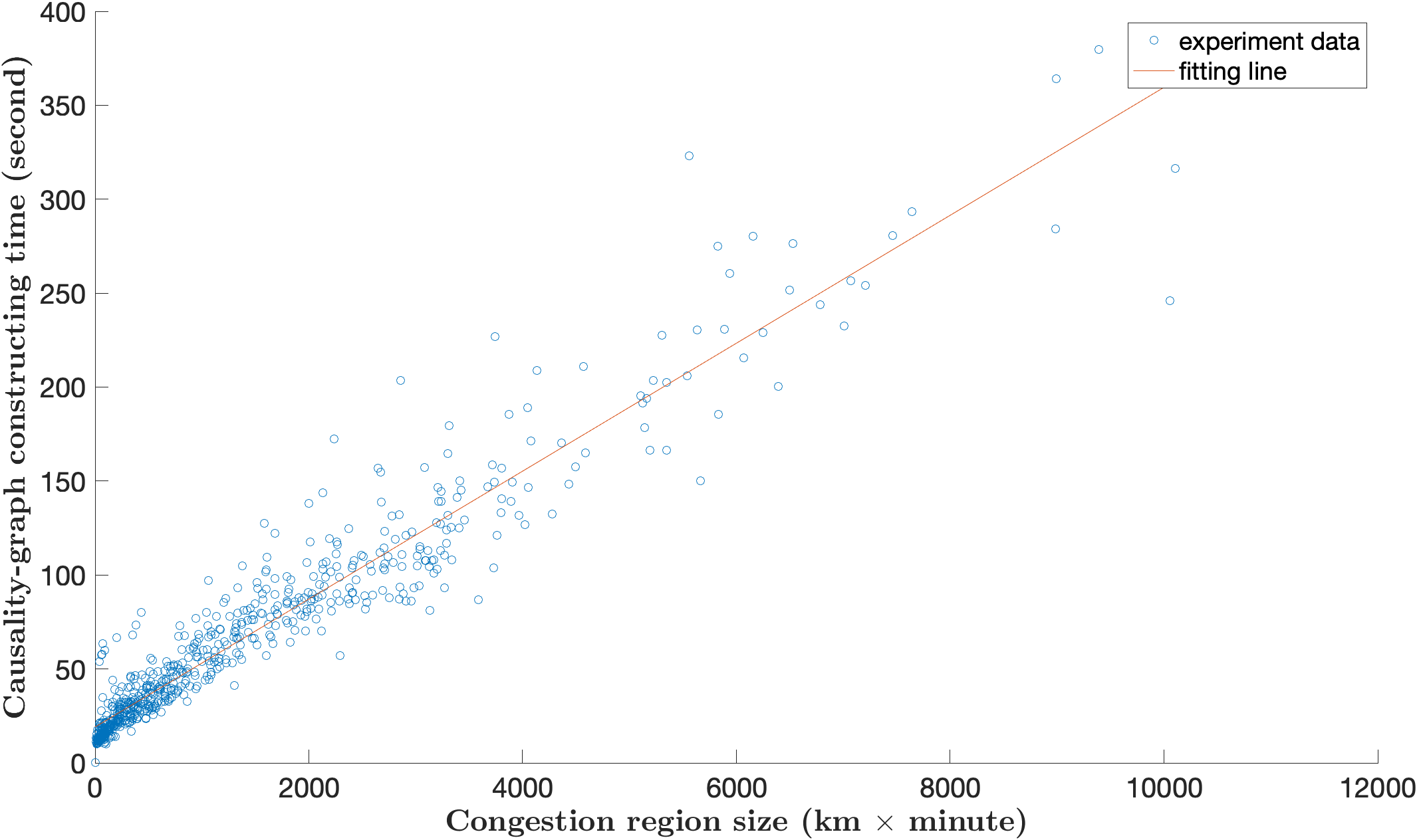}
		\caption{The constructing time of causality-graphs of all the patterns in the experiment data.}
		\label{fig:congsize_segtime}
	\end{minipage}%
\end{figure}

The processing time in the proposed method is spent mainly on relation-graph construction and graph-similarity measurement. Regarding the former, relation graphs of all congestion patterns in the database are pre-processed and registered in advance. Hence, at the moment of retrieval, only the example pattern needs to be parsed. The processing time depends (almost) linearly on the size of the corresponding congestion region (or pattern) as shown in Fig. \ref{fig:congsize_segtime}. In addition, the majority of patterns have sizes of approximately under 1000 (km $\times$ minutes) and take around 60 seconds to build their relation graphs.

\begin{figure}
	\centering
	\begin{minipage}{0.9\textwidth}%
		\centering
		\begin{subfigure}[b]{0.45\textwidth}
			\centering
			\includegraphics[width=\textwidth]{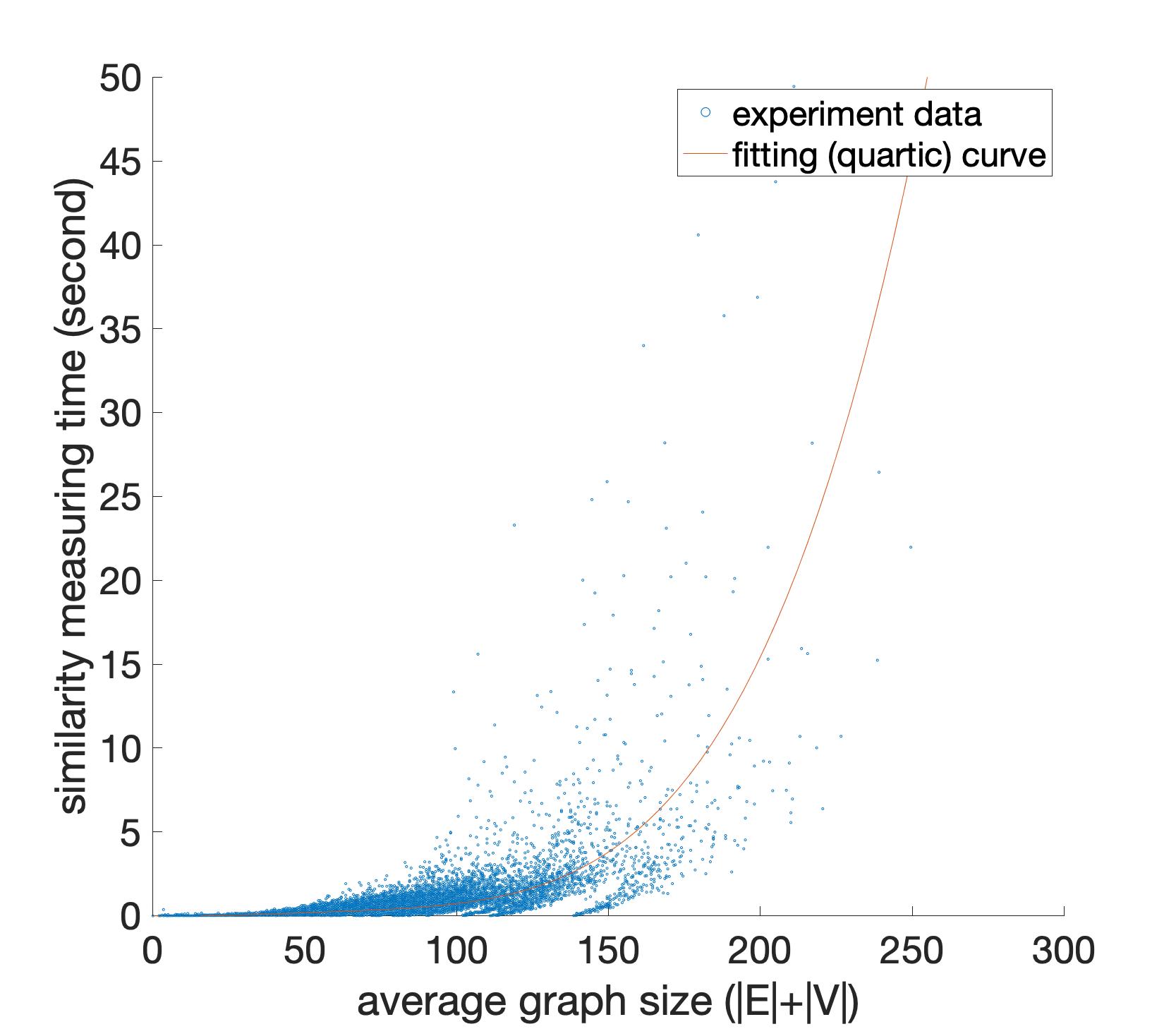}
			\caption{}
			\label{fig:graphsize_simtime_pair}
		\end{subfigure}
		\hfill
		\begin{subfigure}[b]{0.45\textwidth}
			\centering
			\includegraphics[width=\textwidth]{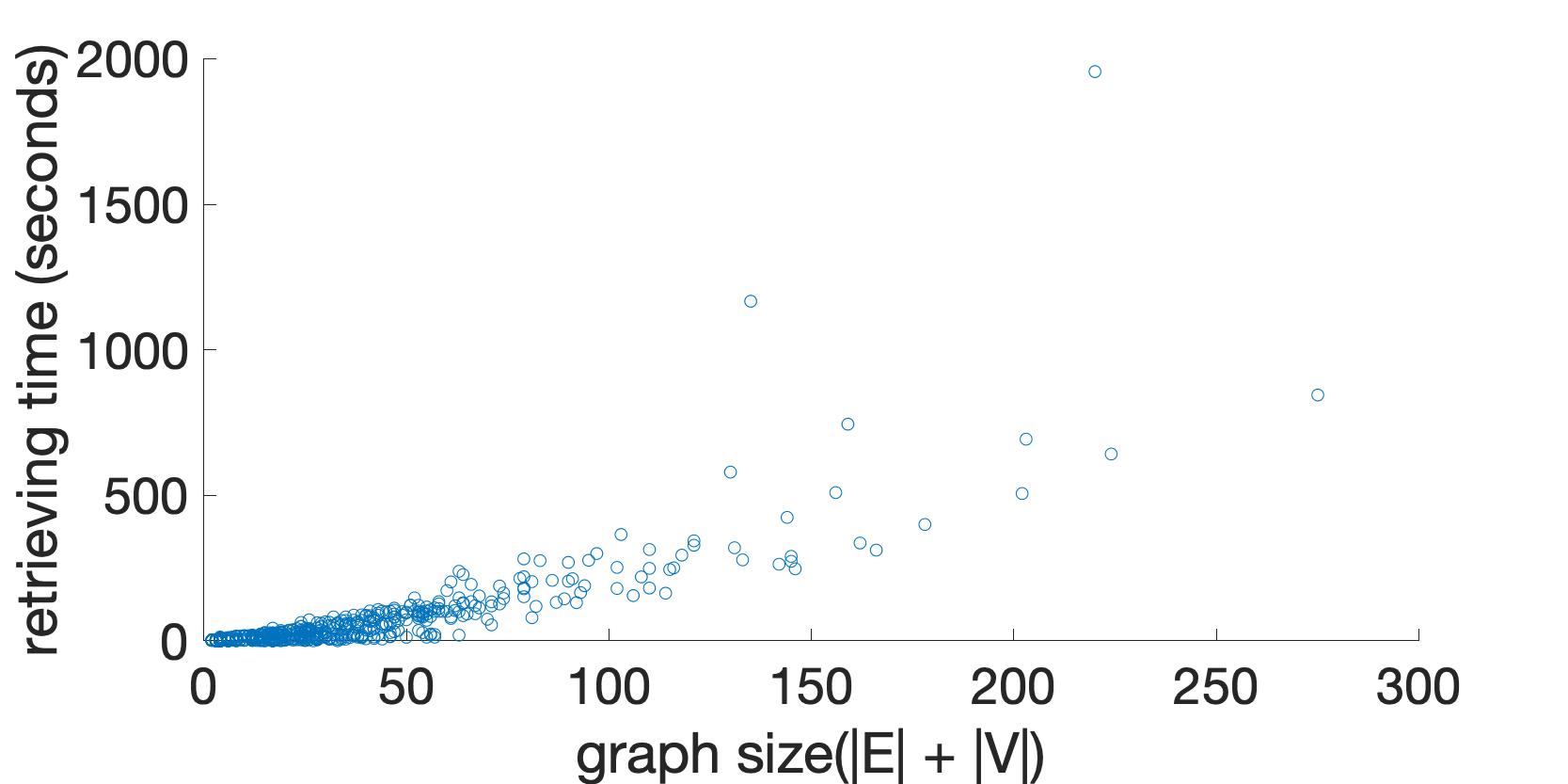}
			\caption{}
			\label{fig:graphsize_simtime_retrieval}
		\end{subfigure}
		\caption{Computation time of measuring the similarity between two relation-graphs: (a) single pair measurement, (b) retrieval time from the whole data (of 778 congestion patterns).}
		\label{fig:graphsize_simtime}
	\end{minipage}%
\end{figure}

Fig. \ref{fig:graphsize_simtime} illustrates the computation of relation-graph similarity. This includes times for matching every single pair as shown in Fig. \ref{fig:graphsize_simtime_pair} and the total retrieving time in the experiment dataset shown in Fig. \ref{fig:graphsize_simtime_retrieval}. It can be expected that the time complexity of the proposed matching method for relation graphs is polynomial w.r.t graph size (measured in the total number of nodes and edges). From a close examination of Fig. \ref{fig:graphsize_simtime_retrieval}, it takes less than one minute to retrieve similar patterns for a pattern of up to 30 nodes plus edges in its relation graph. However, the waiting time can be long for large-scale patterns or a collection of numerous patterns. Therefore, to scale up the proposed method to larger datasets, further improvements are necessary. One approach is to narrow down the search space by some quick pre-processing. For example, as suggested by Fig. \ref{fig:graphsize_simtime_pair}, when retrieving for small-scale patterns, a (computationally) fast filter can be applied to keep only patterns with small numbers of nodes in their relation-graphs. Another approach is to employ more computational power where measuring similarities can be done in parallel processing, hence, reducing the responding time.

\subsection{An opportunity for semantic retrieval}
The relation graph essentially represents traffic patterns in an abstract form that comprises nodes representing traffic phenomena. Therefore, this graph can be used as an alternative representation approach for describing expected patterns for retrieval. This way of searching has the advantage that users have high control of what to expect from the returned patterns, therefore, it is not limited by a given pattern. Take the following expectation as an example: \textit{\textquotedblleft bottleneck that causes upstream congestion with the size of 500 km $\times$ minutes in which multiple disturbances (e.g. 15 instances) emerge\textquotedblright}. A corresponding retrieval can be carried out by constructing a relation graph with two nodes, namely B and D. Their attributes are directly obtained from the description.

\section{Conclusion}
\label{sec_conclusion}

This paper presents a new method for pattern retrieval of highway traffic congestion based on image processing techniques and a graph-matching approach. We demonstrate the efficacy and efficiency of the method on a large-scale traffic database covering the entire Dutch freeway road network over several years.

From the image representations of congestion patterns, traffic-domain elements are extracted by applying image processing techniques. Specifically, those include disturbances (using Watershed-based segmentation), bottlenecks (using gradient filtering based on the direction of characteristic waves), and homogeneous congestion (using texture-based analysis in this paper). 
We propose a so-called relation graph as an abstract representation of the overall congestion patterns with their constituent components (bottlenecks, homogeneous congested regions and disturbances) so that their spatial relations are preserved. We formulate a parameterised matching function to measure the similarity between two relation graphs, reflecting different perspectives on observing patterns, namely component sizes, redundancy penalties, disturbance frequencies, individual completion and structural integrity. 

The case study demonstrates the ability of the proposed method to successfully retrieve complex patterns with various retrieval cases. Most importantly, we show that combining domain knowledge with computer science techniques is highly effective. Using traffic-domain features helps to make query outcomes transparent and easily explainable. Moreover, via the customisable parameters, modifications are made possible to improve a retrieval result according to users' points of view on similarity, i.e. what they are looking for. The relation-graph representation also offers a new opportunity for semantic retrieval, in which expected patterns are described intuitively using some well-known traffic phenomena. This successful development of a retrieval system for 2D congestion patterns essentially motivates various studies in, for example, congestion analysis, and traffic prediction. A broader and more thorough understanding of variations of a type of congestion is beneficial for evaluating mitigation strategies for the corresponding congested traffic. Furthermore, traffic states can be predicted at an abstract level, i.e. pattern level, which in essence is another way to look at traffic at a higher level.

There are open directions for future research to improve the proposed method. From a practical viewpoint, computation time is important. One promising approach is to start with a (rough) classification of the input pattern. Accordingly, the search space is effectively reduced from an entire dataset to the partition representing one single class of patterns. A different approach is a two-step approach that combines generic feature approaches and the proposed approach. Since Euclidean distances are fast to compute, similarities between patterns can be measured quickly based on the former features. Then, the proposed approach only needs to be applied to a reasonable number of top-ranked patterns thereof. Another future research direction is to incorporate more relevant characteristics to the relation graph to refine retrieval outcomes.

%% file: pattern_retrieval.bbl
\begin{thebibliography}{32}
\expandafter\ifx\csname natexlab\endcsname\relax\def\natexlab#1{#1}\fi
\providecommand{\url}[1]{\texttt{#1}}
\providecommand{\href}[2]{#2}
\providecommand{\path}[1]{#1}
\providecommand{\DOIprefix}{doi:}
\providecommand{\ArXivprefix}{arXiv:}
\providecommand{\URLprefix}{URL: }
\providecommand{\Pubmedprefix}{pmid:}
\providecommand{\doi}[1]{\href{http://dx.doi.org/#1}{\path{#1}}}
\providecommand{\Pubmed}[1]{\href{pmid:#1}{\path{#1}}}
\providecommand{\bibinfo}[2]{#2}
\ifx\xfnm\relax \def\xfnm[#1]{\unskip,\space#1}\fi
\bibitem[{ndw()}]{ndw}
, .
\newblock \bibinfo{title}{National datawarehouse of traffic information}.
\newblock \URLprefix \url{http://www.ndw.nu/en/}.
\bibitem[{Bay et~al.(2008)Bay, Ess, Tuytelaars and Van~Gool}]{bay2008speeded}
\bibinfo{author}{Bay, H.}, \bibinfo{author}{Ess, A.},
  \bibinfo{author}{Tuytelaars, T.}, \bibinfo{author}{Van~Gool, L.},
  \bibinfo{year}{2008}.
\newblock \bibinfo{title}{Speeded-up robust features (surf)}.
\newblock \bibinfo{journal}{Computer vision and image understanding}
  \bibinfo{volume}{110}, \bibinfo{pages}{346--359}.
\bibitem[{Blondel et~al.(2004)Blondel, Gajardo, Heymans, Senellart and
  Van~Dooren}]{blondel2004measure}
\bibinfo{author}{Blondel, V.D.}, \bibinfo{author}{Gajardo, A.},
  \bibinfo{author}{Heymans, M.}, \bibinfo{author}{Senellart, P.},
  \bibinfo{author}{Van~Dooren, P.}, \bibinfo{year}{2004}.
\newblock \bibinfo{title}{A measure of similarity between graph vertices:
  Applications to synonym extraction and web searching}.
\newblock \bibinfo{journal}{SIAM review} \bibinfo{volume}{46},
  \bibinfo{pages}{647--666}.
\bibitem[{Bunke(1997)}]{bunke1997relation}
\bibinfo{author}{Bunke, H.}, \bibinfo{year}{1997}.
\newblock \bibinfo{title}{On a relation between graph edit distance and maximum
  common subgraph}.
\newblock \bibinfo{journal}{Pattern Recognition Letters} \bibinfo{volume}{18},
  \bibinfo{pages}{689--694}.
\bibitem[{Calvert et~al.(2018)Calvert, Taale, Snelder and
  Hoogendoorn}]{calvert2018improving}
\bibinfo{author}{Calvert, S.}, \bibinfo{author}{Taale, H.},
  \bibinfo{author}{Snelder, M.}, \bibinfo{author}{Hoogendoorn, S.},
  \bibinfo{year}{2018}.
\newblock \bibinfo{title}{Improving traffic management through consideration of
  uncertainty and stochastics in traffic flow}.
\newblock \bibinfo{journal}{Case Studies on Transport Policy}
  \bibinfo{volume}{6}, \bibinfo{pages}{81--93}.
\bibitem[{Calvert et~al.(2011)Calvert, Van Den~Broek and van
  Noort}]{calvert2011modelling}
\bibinfo{author}{Calvert, S.C.}, \bibinfo{author}{Van Den~Broek, T.A.},
  \bibinfo{author}{van Noort, M.}, \bibinfo{year}{2011}.
\newblock \bibinfo{title}{Modelling cooperative driving in congestion
  shockwaves on a freeway network}, in: \bibinfo{booktitle}{2011 14th
  International IEEE Conference on Intelligent Transportation Systems (ITSC)},
  \bibinfo{organization}{IEEE}. pp. \bibinfo{pages}{614--619}.
\bibitem[{Chan and Vese(2001)}]{chan2001active}
\bibinfo{author}{Chan, T.F.}, \bibinfo{author}{Vese, L.A.},
  \bibinfo{year}{2001}.
\newblock \bibinfo{title}{Active contours without edges}.
\newblock \bibinfo{journal}{IEEE Transactions on image processing}
  \bibinfo{volume}{10}, \bibinfo{pages}{266--277}.
\bibitem[{Conte et~al.(2004)Conte, Foggia, Sansone and Vento}]{conte2004thirty}
\bibinfo{author}{Conte, D.}, \bibinfo{author}{Foggia, P.},
  \bibinfo{author}{Sansone, C.}, \bibinfo{author}{Vento, M.},
  \bibinfo{year}{2004}.
\newblock \bibinfo{title}{Thirty years of graph matching in pattern
  recognition}.
\newblock \bibinfo{journal}{International journal of pattern recognition and
  artificial intelligence} \bibinfo{volume}{18}, \bibinfo{pages}{265--298}.
\bibitem[{Cootes et~al.(1995)Cootes, Taylor, Cooper and
  Graham}]{cootes1995active}
\bibinfo{author}{Cootes, T.F.}, \bibinfo{author}{Taylor, C.J.},
  \bibinfo{author}{Cooper, D.H.}, \bibinfo{author}{Graham, J.},
  \bibinfo{year}{1995}.
\newblock \bibinfo{title}{Active shape models-their training and application}.
\newblock \bibinfo{journal}{Computer vision and image understanding}
  \bibinfo{volume}{61}, \bibinfo{pages}{38--59}.
\bibitem[{Datta et~al.(2008)Datta, Joshi, Li and Wang}]{datta2008image}
\bibinfo{author}{Datta, R.}, \bibinfo{author}{Joshi, D.}, \bibinfo{author}{Li,
  J.}, \bibinfo{author}{Wang, J.Z.}, \bibinfo{year}{2008}.
\newblock \bibinfo{title}{Image retrieval: Ideas, influences, and trends of the
  new age}.
\newblock \bibinfo{journal}{ACM Computing Surveys (Csur)} \bibinfo{volume}{40},
  \bibinfo{pages}{1--60}.
\bibitem[{Emmert-Streib et~al.(2016)Emmert-Streib, Dehmer and
  Shi}]{emmert2016fifty}
\bibinfo{author}{Emmert-Streib, F.}, \bibinfo{author}{Dehmer, M.},
  \bibinfo{author}{Shi, Y.}, \bibinfo{year}{2016}.
\newblock \bibinfo{title}{Fifty years of graph matching, network alignment and
  network comparison}.
\newblock \bibinfo{journal}{Information sciences} \bibinfo{volume}{346},
  \bibinfo{pages}{180--197}.
\bibitem[{Foggia et~al.(2014)Foggia, Percannella and Vento}]{foggia2014graph}
\bibinfo{author}{Foggia, P.}, \bibinfo{author}{Percannella, G.},
  \bibinfo{author}{Vento, M.}, \bibinfo{year}{2014}.
\newblock \bibinfo{title}{Graph matching and learning in pattern recognition in
  the last 10 years}.
\newblock \bibinfo{journal}{International Journal of Pattern Recognition and
  Artificial Intelligence} \bibinfo{volume}{28}, \bibinfo{pages}{1450001}.
\bibitem[{Gao et~al.(2010)Gao, Xiao, Tao and Li}]{gao2010survey}
\bibinfo{author}{Gao, X.}, \bibinfo{author}{Xiao, B.}, \bibinfo{author}{Tao,
  D.}, \bibinfo{author}{Li, X.}, \bibinfo{year}{2010}.
\newblock \bibinfo{title}{A survey of graph edit distance}.
\newblock \bibinfo{journal}{Pattern Analysis and applications}
  \bibinfo{volume}{13}, \bibinfo{pages}{113--129}.
\bibitem[{G{\"a}rtner et~al.(2003)G{\"a}rtner, Flach and
  Wrobel}]{gartner2003graph}
\bibinfo{author}{G{\"a}rtner, T.}, \bibinfo{author}{Flach, P.},
  \bibinfo{author}{Wrobel, S.}, \bibinfo{year}{2003}.
\newblock \bibinfo{title}{On graph kernels: Hardness results and efficient
  alternatives}, in: \bibinfo{booktitle}{Learning theory and kernel machines}.
  \bibinfo{publisher}{Springer}, pp. \bibinfo{pages}{129--143}.
\bibitem[{Haralick et~al.(1973)Haralick, Shanmugam and
  Dinstein}]{haralick1973textural}
\bibinfo{author}{Haralick, R.M.}, \bibinfo{author}{Shanmugam, K.},
  \bibinfo{author}{Dinstein, I.H.}, \bibinfo{year}{1973}.
\newblock \bibinfo{title}{Textural features for image classification}.
\newblock \bibinfo{journal}{IEEE Transactions on systems, man, and cybernetics}
  , \bibinfo{pages}{610--621}.
\bibitem[{Helbing et~al.(2009)Helbing, Treiber, Kesting and
  Sch{\"o}nhof}]{helbing2009theoretical}
\bibinfo{author}{Helbing, D.}, \bibinfo{author}{Treiber, M.},
  \bibinfo{author}{Kesting, A.}, \bibinfo{author}{Sch{\"o}nhof, M.},
  \bibinfo{year}{2009}.
\newblock \bibinfo{title}{Theoretical vs. empirical classification and
  prediction of congested traffic states}.
\newblock \bibinfo{journal}{The European Physical Journal B-Condensed Matter
  and Complex Systems} \bibinfo{volume}{69}, \bibinfo{pages}{583--598}.
\bibitem[{Krishnakumari et~al.(2017)Krishnakumari, Nguyen, Heydenrijk-Ottens,
  Vu and van Lint}]{krishnakumari2017traffic}
\bibinfo{author}{Krishnakumari, P.}, \bibinfo{author}{Nguyen, T.},
  \bibinfo{author}{Heydenrijk-Ottens, L.}, \bibinfo{author}{Vu, H.L.},
  \bibinfo{author}{van Lint, H.}, \bibinfo{year}{2017}.
\newblock \bibinfo{title}{Traffic congestion pattern classification using
  multiclass active shape models}.
\newblock \bibinfo{journal}{Transportation Research Record: Journal of the
  Transportation Research Board} , \bibinfo{pages}{94--103}.
\bibitem[{Kuhn(1955)}]{kuhn1955hungarian}
\bibinfo{author}{Kuhn, H.W.}, \bibinfo{year}{1955}.
\newblock \bibinfo{title}{The hungarian method for the assignment problem}.
\newblock \bibinfo{journal}{Naval research logistics quarterly}
  \bibinfo{volume}{2}, \bibinfo{pages}{83--97}.
\bibitem[{Lopez et~al.(2017)Lopez, Leclercq, Krishnakumari, Chiabaut and van
  Lint}]{lopez2017revealing}
\bibinfo{author}{Lopez, C.}, \bibinfo{author}{Leclercq, L.},
  \bibinfo{author}{Krishnakumari, P.}, \bibinfo{author}{Chiabaut, N.},
  \bibinfo{author}{van Lint, H.}, \bibinfo{year}{2017}.
\newblock \bibinfo{title}{Revealing the day-to-day regularity of urban
  congestion patterns with 3d speed maps}.
\newblock \bibinfo{journal}{Scientific Reports} \bibinfo{volume}{7},
  \bibinfo{pages}{1--11}.
\bibitem[{Nguyen et~al.(2016)Nguyen, Krishnakumari, Vu and van
  Lint}]{nguyen2016traffic}
\bibinfo{author}{Nguyen, H.N.}, \bibinfo{author}{Krishnakumari, P.},
  \bibinfo{author}{Vu, H.L.}, \bibinfo{author}{van Lint, H.},
  \bibinfo{year}{2016}.
\newblock \bibinfo{title}{Traffic congestion pattern classification using
  multi-class svm}, in: \bibinfo{booktitle}{Intelligent Transportation Systems
  (ITSC), 2016 IEEE 19th International Conference on},
  \bibinfo{organization}{IEEE}. pp. \bibinfo{pages}{1059--1064}.
\bibitem[{Nguyen et~al.(2021)Nguyen, Calvert, Vu and van
  Lint}]{nguyen2021automated}
\bibinfo{author}{Nguyen, T.T.}, \bibinfo{author}{Calvert, S.C.},
  \bibinfo{author}{Vu, H.L.}, \bibinfo{author}{van Lint, H.},
  \bibinfo{year}{2021}.
\newblock \bibinfo{title}{An automated detection framework for multiple highway
  bottleneck activations}.
\newblock \bibinfo{journal}{IEEE Transactions on Intelligent Transportation
  Systems} .
\bibitem[{Nguyen et~al.(2019)Nguyen, Krishnakumari, Calvert, Vu and
  Van~Lint}]{nguyen2019feature}
\bibinfo{author}{Nguyen, T.T.}, \bibinfo{author}{Krishnakumari, P.},
  \bibinfo{author}{Calvert, S.C.}, \bibinfo{author}{Vu, H.L.},
  \bibinfo{author}{Van~Lint, H.}, \bibinfo{year}{2019}.
\newblock \bibinfo{title}{Feature extraction and clustering analysis of highway
  congestion}.
\newblock \bibinfo{journal}{Transportation Research Part C: Emerging
  Technologies} \bibinfo{volume}{100}, \bibinfo{pages}{238--258}.
\bibitem[{Riesen(2015)}]{riesen2015structural}
\bibinfo{author}{Riesen, K.}, \bibinfo{year}{2015}.
\newblock \bibinfo{title}{Structural pattern recognition with graph edit
  distance}, in: \bibinfo{booktitle}{Advances in computer vision and pattern
  recognition}. \bibinfo{publisher}{Springer}.
\bibitem[{Schreiter et~al.(2010)Schreiter, van Lint, Treiber and
  Hoogendoorn}]{schreiter2010two}
\bibinfo{author}{Schreiter, T.}, \bibinfo{author}{van Lint, H.},
  \bibinfo{author}{Treiber, M.}, \bibinfo{author}{Hoogendoorn, S.},
  \bibinfo{year}{2010}.
\newblock \bibinfo{title}{Two fast implementations of the adaptive smoothing
  method used in highway traffic state estimation}, in:
  \bibinfo{booktitle}{Intelligent Transportation Systems (ITSC), 2010 13th
  International IEEE Conference on}, \bibinfo{organization}{IEEE}. pp.
  \bibinfo{pages}{1202--1208}.
\bibitem[{Soriguera and Robust{\'e}(2011)}]{soriguera2011estimation}
\bibinfo{author}{Soriguera, F.}, \bibinfo{author}{Robust{\'e}, F.},
  \bibinfo{year}{2011}.
\newblock \bibinfo{title}{Estimation of traffic stream space mean speed from
  time aggregations of double loop detector data}.
\newblock \bibinfo{journal}{Transportation research part C: emerging
  technologies} \bibinfo{volume}{19}, \bibinfo{pages}{115--129}.
\bibitem[{Treiber and Helbing(2002)}]{treiber2002reconstructing}
\bibinfo{author}{Treiber, M.}, \bibinfo{author}{Helbing, D.},
  \bibinfo{year}{2002}.
\newblock \bibinfo{title}{Reconstructing the spatio-temporal traffic dynamics
  from stationary detector data}.
\newblock \bibinfo{journal}{Cooperative Transportation Dynamics}
  \bibinfo{volume}{1}, \bibinfo{pages}{3--1}.
\bibitem[{Van~Lint et~al.(2005)Van~Lint, Hoogendoorn and van
  Zuylen}]{van2005accurate}
\bibinfo{author}{Van~Lint, J.}, \bibinfo{author}{Hoogendoorn, S.},
  \bibinfo{author}{van Zuylen, H.J.}, \bibinfo{year}{2005}.
\newblock \bibinfo{title}{Accurate freeway travel time prediction with
  state-space neural networks under missing data}.
\newblock \bibinfo{journal}{Transportation Research Part C: Emerging
  Technologies} \bibinfo{volume}{13}, \bibinfo{pages}{347--369}.
\bibitem[{Vlahogianni et~al.(2005)Vlahogianni, Karlaftis and
  Golias}]{vlahogianni2005optimized}
\bibinfo{author}{Vlahogianni, E.I.}, \bibinfo{author}{Karlaftis, M.G.},
  \bibinfo{author}{Golias, J.C.}, \bibinfo{year}{2005}.
\newblock \bibinfo{title}{Optimized and meta-optimized neural networks for
  short-term traffic flow prediction: A genetic approach}.
\newblock \bibinfo{journal}{Transportation Research Part C: Emerging
  Technologies} \bibinfo{volume}{13}, \bibinfo{pages}{211--234}.
\bibitem[{Wang et~al.(2006)Wang, Papageorgiou and
  Messmer}]{wang2006renaissance}
\bibinfo{author}{Wang, Y.}, \bibinfo{author}{Papageorgiou, M.},
  \bibinfo{author}{Messmer, A.}, \bibinfo{year}{2006}.
\newblock \bibinfo{title}{Renaissance--a unified macroscopic model-based
  approach to real-time freeway network traffic surveillance}.
\newblock \bibinfo{journal}{Transportation Research Part C: Emerging
  Technologies} \bibinfo{volume}{14}, \bibinfo{pages}{190--212}.
\bibitem[{van~de Weg et~al.(2018)van~de Weg, Vu, Hegyi and
  Hoogendoorn}]{van2018hierarchical}
\bibinfo{author}{van~de Weg, G.S.}, \bibinfo{author}{Vu, H.L.},
  \bibinfo{author}{Hegyi, A.}, \bibinfo{author}{Hoogendoorn, S.P.},
  \bibinfo{year}{2018}.
\newblock \bibinfo{title}{A hierarchical control framework for coordination of
  intersection signal timings in all traffic regimes}.
\newblock \bibinfo{journal}{IEEE Transactions on Intelligent Transportation
  Systems} \bibinfo{volume}{20}, \bibinfo{pages}{1815--1827}.
\bibitem[{Zager and Verghese(2008)}]{zager2008graph}
\bibinfo{author}{Zager, L.A.}, \bibinfo{author}{Verghese, G.C.},
  \bibinfo{year}{2008}.
\newblock \bibinfo{title}{Graph similarity scoring and matching}.
\newblock \bibinfo{journal}{Applied mathematics letters} \bibinfo{volume}{21},
  \bibinfo{pages}{86--94}.
\bibitem[{Zhou et~al.(2017)Zhou, Li and Tian}]{zhou2017recent}
\bibinfo{author}{Zhou, W.}, \bibinfo{author}{Li, H.}, \bibinfo{author}{Tian,
  Q.}, \bibinfo{year}{2017}.
\newblock \bibinfo{title}{Recent advance in content-based image retrieval: A
  literature survey}.
\newblock \bibinfo{journal}{arXiv preprint arXiv:1706.06064} .

\end{thebibliography}
